%% file: learning-on-hardware.tex
\documentclass[acmsmall]{acmart}

\setcopyright{rightsretained}
\copyrightyear{2021}
\acmYear{2021}
\acmDOI{x}
\acmJournal{arXiv}
\acmVolume{0}
\acmNumber{0}
\acmArticle{0}
\acmMonth{4}

\usepackage[colorinlistoftodos]{todonotes} 
\usepackage{soul} 

\usepackage{amsmath,amsfonts}
\usepackage{algorithmic}
\usepackage{graphicx}
\usepackage{textcomp}
\usepackage{xcolor}
\usepackage[acronym,nonumberlist,nowarn,style=long]{glossaries}
\usepackage{amsmath}
\newcommand{\done}[1]{}
\usepackage{subcaption }
\usepackage{colortbl}
\usepackage{array}
\usepackage{svg}
\usepackage{float}
\usepackage{pdflscape}
\usepackage[acronym,nonumberlist,nowarn,style=long]{glossaries}
\usepackage{csquotes}
\usepackage{siunitx}
\usepackage{booktabs}
\usepackage{tabularx}
\newcolumntype{C}{>{\centering\arraybackslash}X} 
\usepackage{multirow}
\usepackage{pdflscape}
\usepackage{afterpage}

\usepackage{cleveref}
\usepackage{url}
\newglossary[slg]{symbolslist}{syi}{syg}{Symbols}
\input{Acronyms.tex}
\input{Symbols.tex}

\begin{document}

\title{Learning on Hardware: A Tutorial on \\ Neural Network Accelerators and Co-Processors
}

\author{Lukas Baischer}
\email{lukas.baischer@student.tuwien.ac.at}
\author{Matthias Wess}
\email{matthias.wess@student.tuwien.ac.at}
\author{Nima TaheriNejad}
\email{nima.taherinejad@tuwien.ac.at}

\affiliation{%
  \institution{TU Wien, Institute of Computer Technology}
  \streetaddress{Gu{\ss}hausstra{\ss}e 27-29/384}
  \city{Vienna}
  \state{Vienna}
  \country{Austria}
  \postcode{1040}
}
\begin{abstract}
\Glspl{DNN} have the advantage that they can take into account a large number of parameters, which enables them to solve complex tasks. In computer vision and speech recognition, they have a better accuracy than common algorithms, and in some tasks, they boast an even higher accuracy than human experts. With the progress of \glspl{DNN} in recent years, many other fields of application such as diagnosis of diseases and autonomous driving are taking advantage of them. 
The trend at \glspl{DNN} is clear: The network size is growing exponentially, which leads to an exponential increase in computational effort and required memory size.
For this reason, optimized hardware accelerators are used to increase the performance of the inference of neuronal networks. However, there are various neural network hardware accelerator platforms, such as \glspl{GPU}, \glspl{ASIC} and \glspl{FPGA}. Each of these platforms offer certain advantages and disadvantages. Also, there are various methods for reducing the computational effort of \glspl{DNN}, which are differently suitable for each hardware accelerator. In this article an overview of existing neural network hardware accelerators and acceleration methods is given. Their strengths and weaknesses are shown and a recommendation of suitable applications is given. In particular, we focus on acceleration of the inference of \glspl{CNN} used for image recognition tasks. Given that there exist many different hardware architectures. \Gls{FPGA}-based implementations are well-suited to show the effect of \gls{DNN} optimization methods on accuracy and throughput. For this reason, the focus of this work is more on \gls{FPGA}-based implementations.
\end{abstract}


\maketitle

\section{Introduction}
Deep learning is currently one of the most prominent machine learning approaches for solving complex tasks that could only be solved by human beforehand \cite{zhang2018survey}. In applications such as computer vision or speech recognition, \glspl{DNN} achieve higher accuracy compared to non-learning algorithms and in some cases even higher than human experts \cite{Sze.2017,Xu.2018,zhang2018survey}. 
The higher accuracy of \glspl{DNN} compared to non-learning algorithms comes from the ability to extract high-level features from the input data after using statistical learning over a high number of training data. Statistical learning leads to an efficient representation of the input space and a good generalization \cite{Sze.2017}. \\
However, this capability requires high computational effort \cite{Sze.2017}. It has shown that, by increasing the number of parameters, the accuracy of a network can be increased \cite{Xu.2018}. Consequently, the trend in \glspl{DNN} is clearly that the network size is growing exponentially. Which leads to an exponentially increasing computational effort and required memory size \cite{Xu.2018}. Therefore, \glspl{CPU} can hardly handle the necessary computations. Hence, structurally optimized hardware accelerators are used to increase the inference performance of neuronal networks. For inference of a neural network running on edge devices, energy efficiency is an important factor that has to be considered, in addition to throughput \cite{Xu.2018}. 

In this paper, we focus on optimization methods and hardware accelerators used for inference of \glspl{CNN}. \Gls{FPGA}-based implementations are discussed in more detail, since they are well suited to show different architecture approaches and how optimization methods affect throughput and accuracy. However, in addition to hardware accelerators mentioned in this paper, other promising approaches such as \gls{SNN} or \gls{IMC} exist. Nevertheless, these approaches require different optimization methods or different math, which is beyond the scope of this paper. Below is an outline of the paper.

\Cref{sec:nn-back} gives a general introduction to the field of neural networks. The intention is to explain the necessary background to understand how neural networks work and to show the basic mathematical operations. \Cref{sec:gaps} highlights the need for neural network hardware accelerators and shows why simply using general-purpose \glspl{GPU} cannot fulfill future requirements in terms of performance density and energy efficiency. Based on the knowledge that more efficient neural network structures are required to meet future requirements, \Cref{sec:methods-dnn-size} shows methods for decreasing the computation effort, required for deep neural networks. In \Cref{sec:algorithmic} algorithmic optimization methods are explained, which increase the algorithmic efficiency and therefore, increase the throughput of hardware accelerators. \Cref{sec:LoopOptimization} explains the parallelization strategies used by neural network hardware accelerators. \Cref{sec:gpu}, \Cref{sec:asic} and \Cref{sec:fpga} present the most commonly used neural network hardware accelerator platforms, namely \glspl{GPU}, \glspl{ASIC} and \glspl{FPGA}. Additionally, an overview of existing implementations is introduced. \Cref{sec:comparision} compares the platforms in terms of speed, accuracy, power, and usability. Finally, we draw our conclusions in \cref{sec_conclusions}.


\section{Neural Networks} \label{sec:nn-back}
\begin{figure}[h]
	\centering
	\includesvg[width=0.5\textwidth]{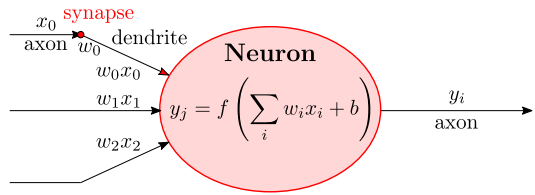}
	\caption[A single neuron and the connection to human brain.] {A single neuron and the connection to human brain. $x_i$, $w_i$, $f(\cdot)$, $b$ and $y_j$ are the input activations, weights, non-linear activation function, bias and output activation. Adapted from \cite[Fig. 2]{Sze.2017}.} 
	\label{fig:neuron-ex}
\end{figure}

Neural networks are a part of the broad field of \gls{AI}. \Gls{AI} generally deals with building intelligent machines, which can solve tasks similar to human \cite{Sze.2017}. Neural networks are inspired by the human brain. \Cref{fig:neuron-ex} visualizes the principle of a single neuron in a neural network and the similarities to the human brain. Additionally, the fundamental \Cref{eq:neuron} for computing the output activation of a single neuron is depicted. $x_i$, $w_i$, $f(\cdot)$, $b$ and $y_j$ are the input activations, weights, non-linear activation function, bias and output activation \cite{Sze.2017}.   \\
\begin{equation}
	y_j = f \left( \sum_i w_i x_i+b\right)
	\label{eq:neuron}
\end{equation}
In principle, any non-linear function can is allowed as an activation function for a neural network. Historically the Sigmoid function was the most popular activation function. However in modern \glspl{DNN} mostly \gls{Relu} is used \cite{Sze.2017}. It has shown that \gls{Relu} achieves good results and provides the benefit of less computational effort since only negative values are to zero. All others stay untouched \cite{AlexKrizhevsky.2012}. However, cutting off all negative neuron outputs leads to a loss of information. For this reason leaky \gls{Relu} is used in order to additionally consider negative neuron outputs \cite{bochkovskiy2020yolov4}.\\
\\\
In a neuronal network, single neurons are arranged in layers. How neurons are arranged in a layer and how they are connected to the successor layer indicates the type of layer or network. Three important neural network types are: 
\glsreset{FC}
\glsreset{CNN}
\glsreset{RNN}
\begin{itemize}
    \item \Gls{FC} or feed-forward neural networks
    \item \Glspl{RNN}
    \item \Glspl{CNN}
\end{itemize}
Fully connected neural networks are historically the first form. In modern networks, they are often used as part of \glspl{CNN} or \glspl{RNN}. In some literature \gls{FC}-layers are called a dense layer, which comes from the high parameter density of fully connected neural networks. \Glspl{CNN} are mainly used in the field of computer vision since \glspl{CNN} consider geometric information. Additionally, \glspl{CNN} require less memory space, since weights in a filter stay constant for a single layer. \Glspl{RNN} are mainly used in tasks such as speech recognition or predictions in the financial world because they provide the possibility to consider the information of former events. This is done by using feedback from a layer to a previous one.  \gls{DNN}, on the other hand, do not necessarily designate the structure of a network. It is related to the keyword deep learning, which should indicate that particularly deep networks, i.e. networks with many layers, are used \cite{Sze.2017}. 
\subsection{Fully connected network}
\begin{figure}
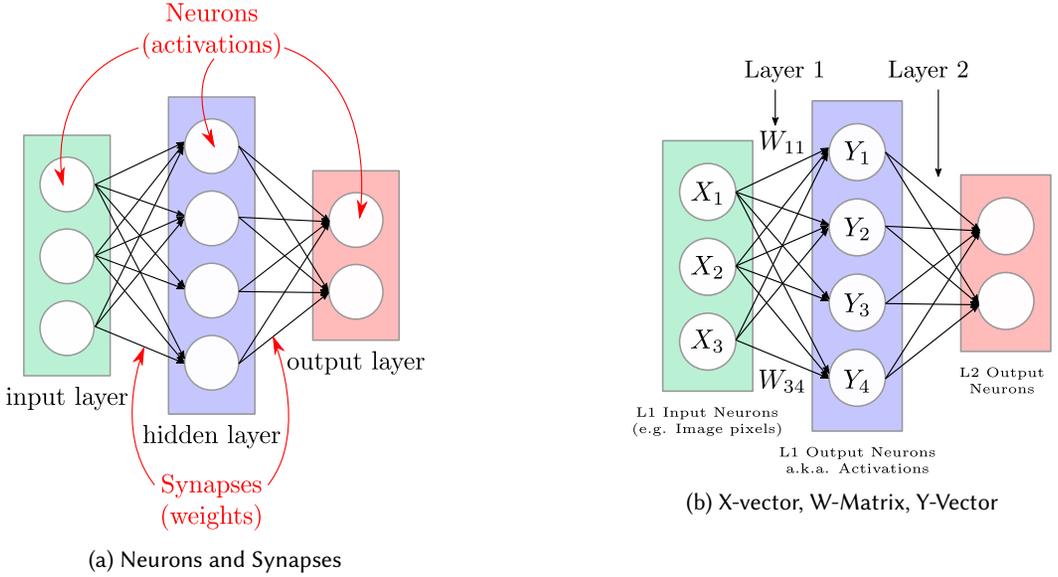

	\centering
	\begin{subfigure}[c]{0.40\textwidth}
		\includesvg[width=\textwidth]{graphics/fc-layer-ex.svg}
		\caption{Neurons and Synapses}
		\label{subfig:fc-layer-ex}
	\end{subfigure}%
	\hfill
	\begin{subfigure}[c]{0.40\textwidth}
		\includesvg[width=\textwidth]{graphics/fc-layer-xwy.svg}
		\caption{X-vector, W-Matrix, Y-Vector}
		\label{subfig:fc-layer-xwy}
	\end{subfigure}%
	\caption[Simple neural network example and terminology.]{Simple neural network example and terminology. (\subref{subfig:fc-layer-ex}) shows a simple example of a fully connected neural network. (\subref{subfig:fc-layer-xwy}) gives an overview how the X-Vector, W-matrix, Y-vector are interpreted. Adapted from \cite[Fig. 3]{Sze.2017}.} 
	\label{fig:nn-example}
\end{figure}

\Cref{fig:nn-example} visualizes the principle and terminology of a fully connected network. The first layer is called the input layer, which receives the data that contains the information to be analyzed. That can be, for example, an image or a sampled audio sequence. The layers which are processing the data are called hidden layers. The output layer is the last layer of a neural network and provides the results of a neural network. Usually, the output of a neural network is represented in the form of a probability of specific possible results. Each layer of a fully connected network is described by \Cref{eq:neuron}, whereas the weight matrix $w_i$ contains an entry for each neuron connection between two layers.

\subsection{CNN}
\begin{figure}[h]
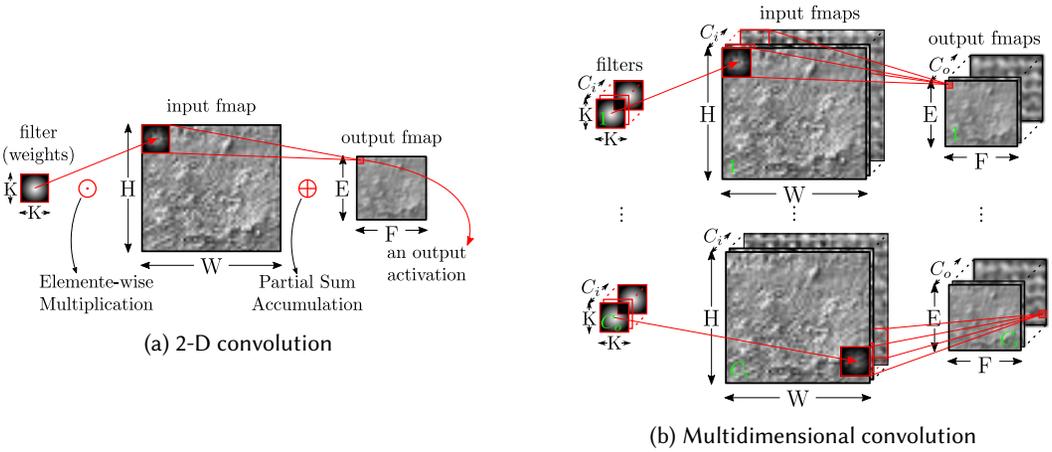

	\centering
	\begin{subfigure}[c]{0.45\textwidth}
		\includesvg[width=\textwidth]{graphics/conv-basic.svg}
		\caption{2-D convolution}
		\label{subfig:conv-basic}
	\end{subfigure}%
	\hfill
	\begin{subfigure}[c]{0.45\textwidth}
		\includesvg[width=\textwidth]{graphics/conv-channels.svg}
		\caption{Multidimensional convolution}
		\label{subfig:conv-channel}
	\end{subfigure}%
	\caption[Convolutional neural network principle.]{Convolutional neural network principle. (\subref{subfig:conv-basic}) shows the principle of a 2-D convolution used in image processing. (\subref{subfig:conv-channel}) visualizes the principle of convolutions in high dimensional \glspl{CNN}. Adapted from \cite[Fig. 9]{Sze.2017}.} 
	\label{fig:conv-example}
\end{figure}

\glsreset{CNN} 
\Glspl{CNN} are state of the art for images classification and detection tasks. Compared to other neural network structures, \glspl{CNN} can consider geometric information included in each pixel, which enables them to surpass the accuracy of fully connected networks using fewer parameters \cite{AlexKrizhevsky.2012}. The 2-D convolution operation is fundamental for \glspl{CNN}. A filter with size $k\times k$ moves along an image or feature map producing a single output image at each step. This is visualized by \Cref{subfig:conv-basic}. It shows that at each step an element-wise multiplication of the filter with the overlapping image tile is performed. The results of the element-wise multiplication are accumulated. \Cref{eq:conv2d} shows how to compute a single output pixel using a 2-D convolution \cite{Abdelouahab.26052018,Sze.2017,Shawahna.2019}.  \\
\begin{equation}
conv2D \left( X^{conv}[c],\Theta[n,c] \right) = \sum X^{conv}[c] \odot \Theta[n,c] 
\label{eq:conv2d}
\end{equation}
Multiple filters are applied in parallel to create complex high-level feature maps for object classification. \Cref{subfig:conv-channel} visualizes that $C_o$ filter applied on a single input image create $C_o$ output feature maps. Parallel feature maps in a single layer are called \textit{channels}. For computing a single pixel of a feature map, the results of the 2-D convolution of all $C_i$ input channels are accumulated. After the summation a non-linear function $f(\cdot)$ is applied. The weights of each filter are different for each channel. That leads to $C_i \times C_o$ different filter for a single layer. \Cref{eq:conv} shows how to compute a single output pixel of a multidimensional convolutional layer \cite{Abdelouahab.26052018}. 
\begin{equation}
Y^{conv}[n] = f\left(\sum_{c=0}^{C} conv2D \left( X^{conv}[c],\Theta[n,c] \right)\right)
\label{eq:conv}
\end{equation}

It visualizes that for computing an output feature map of a single \gls{CNN} layer, $C_o \cdot C_i \cdot W \cdot H \cdot K \cdot K$ \gls{MAC} operations are required. The huge computational effort, which is necessary for \glspl{CNN}, leads to the need of dedicated and structural optimized hardware acceleration platforms such as \glspl{GPU}, \glspl{FPGA} and \glspl{ASIC} \cite{Abdelouahab.26052018}.   

\subsection{RNN}
\glsreset{RNN} 
In contrast to \gls{FC}-neural networks and \glspl{CNN}, \glspl{RNN} use intralayer recurrent connection. These recurrent connections allow \glspl{RNN} to consider past results or states in addition to the current input data \cite{deng2020model,Sze.2017}. Due to this ability, \glspl{RNN} are particularly suitable for tasks like speech recognition, translation, and financial predictions. \\ 
\begin{figure}[h]
	\centering
	\includesvg[width=0.35\textwidth]{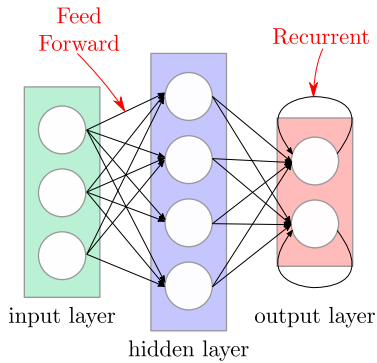}
	\caption[Recurrent neural network.] {Simplified structure of a recurrent neural network \cite[Fig. 8]{Sze.2017}.} 
	\label{fig:rnn}
\end{figure} 

\Cref{fig:rnn} shows a simplified structure of a \gls{RNN}. It visualizes that the output of the networks is traced back to the input of the output layer. That enables a \gls{RNN} to additionally consider the former output of the network. Recurrent layer are mostly implemented using \gls{LSTM} or \gls{GRU} layer \cite{deng2020model}.

\section{Gaps between accuracy requirement and hardware performance} \label{sec:gaps}
\begin{figure}
	\centering
	\begin{subfigure}[c]{\textwidth}
		\includegraphics[width=\textwidth]{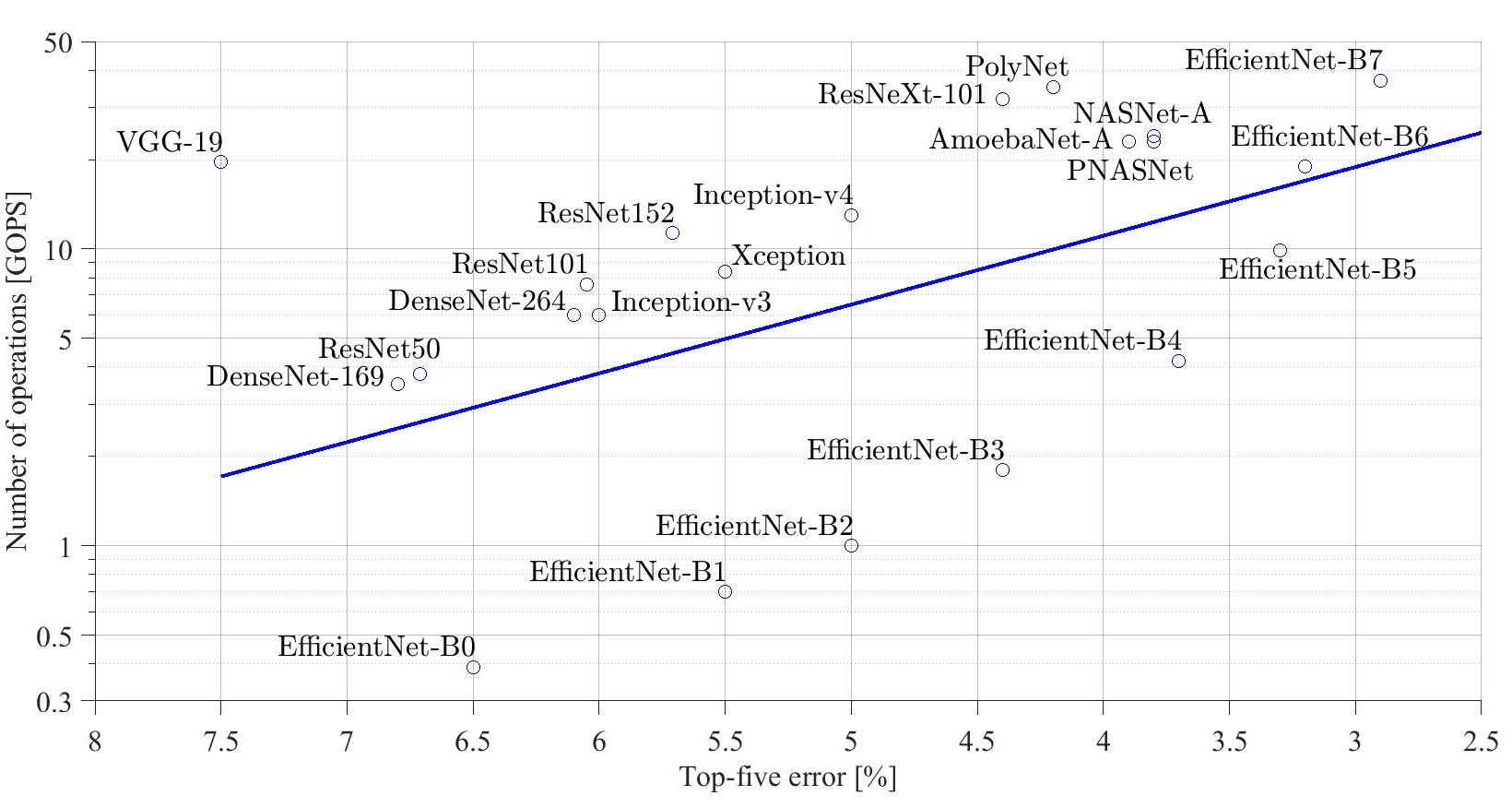}
	    \caption{}
		\label{subfig:gap-nn_exp}
	\end{subfigure}%
	\hfill
	\begin{subfigure}[c]{\textwidth}
		\includegraphics[width=\textwidth]{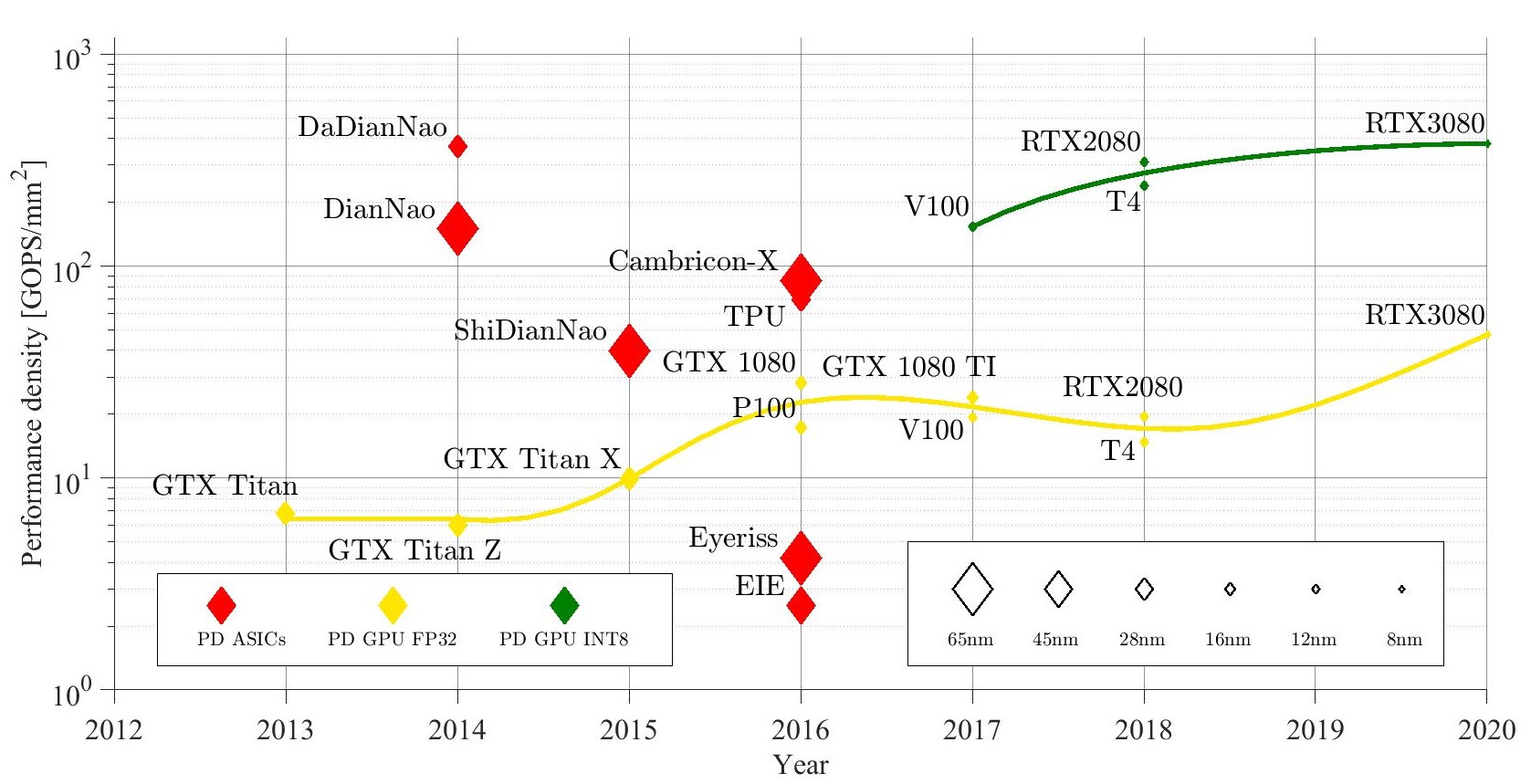}
		\caption{}
		\label{subfig:gap-pd}
	\end{subfigure}%
	\caption[Gap between required number of operations and performance density]{Gap between required number of operations and performance density. (\subref{subfig:gap-nn_exp}) Number of operations versus top-five error rate for leading DNN designs from ImageNet classification competition. Data from: \cite{he2016deep,tan2019efficientnet}. (\subref{subfig:gap-pd}) Performance density (PD) of leading GPU and ASIC platforms. To catch up with the required number of operations, simply increasing the chip area is not feasible. Data from \cite{Chen.2016c,chen2014dadiannao,Han.1806201622062016,zhang2016cambricon,Chen.2017,jouppi2017datacenter,NVIDIA.2017,NVIDIA.turing,nvidia.ampere,Xu.2018}, Adopted from \cite[Fig. 3]{Xu.2018}} 
    \label{fig:DNN-gap}
\end{figure} 

\Cref{subfig:gap-nn_exp} shows the required number of operations for structural optimized neural networks versus the achievable top-5 error using the ImageNet dataset. The blue line visualizes the for a linear increase of the accuracy leads to an exponential increase of the computational effort. \Cref{subfig:gap-pd} visualizes the improvement of the performance density of neural network hardware accelerators. Comparing \Cref{subfig:gap-nn_exp} and \Cref{subfig:gap-pd} concludes that the number of operation required to compute a neural network increases faster than the hardware performance can follow. That leads to a gap between the required hardware performance and the actual available hardware performance \cite{Xu.2018}. \Cref{subfig:gap-pd} shows that the performance density of \glspl{GPU} using 32bit floating-point operations depends mainly on the feature size\footnote{In this context, feature size refers to the technology used to manufacture the \gls{IC} and not on the number of feature maps in a CNN} of the manufactured die. The feature size is visualized by the size of the diamonds in \Cref{subfig:gap-pd}. It is also shown that a simple way to increase the performance density of a hardware accelerator is to reduce the bit width of arithmetic operations. That is the main reason why \glspl{ASIC} and \glspl{FPGA} can reach a higher performance density compared to general-purpose \glspl{GPU} \cite{Xu.2018}. For this reason, modern \glspl{GPU} implement specialized fixed-point arithmetic units, which increases the performance density and power efficiency of \glspl{GPU} \cite{NVIDIA.turing,nvidia.ampere}. However, \glspl{GPU} still require additional 32-bit floating-point units for use as graphics cards. Nevertheless, \glspl{GPU} exhibit a higher throughput compared to \glspl{FPGA} and most \glspl{ASIC} \cite{Xu.2018,Sze.2017}. \\
The performance and energy efficiency of hardware accelerators is tied to the feature size of the respective manufacturing technology. However, the feature size of a die can only be decreased to a certain point due to Moore's law end. Therefore, increasing the performance density at the same speed as the network parameter number is not possible \cite{Xu.2018}.


\section{DNN compression methods} \label{sec:methods-dnn-size} 
To narrow the gap between accuracy requirement and hardware performance, substantially two approaches are followed. On the one hand, compact models and pruning create more efficient neural network structures. On the other hand, decreasing the bit width of weights and activations leads to more efficient hardware \cite{Wang.2019,Xu.2018,Sze.2017}. General-purpose \glspl{GPU} are usually designed for computation of large floating-point vectors. Since sparsity reduces the vector size and quantization focus on fixed-point operations, general-purpose \glspl{GPU} have little profit from these methods. Due to their greater freedom of design, \gls{ASIC} and \gls{FPGA} based hardware accelerators achieve higher efficiency for such structures \cite{Wang.2019,Xu.2018,Sze.2017}. \\
\\\
\Glspl{SNN} use discreet spikes that propagate through the network instead of activations represented as a number. They are a more biologically realistic replica of the human brain since biological neurons use a similar technique. \Glspl{SNN} bring the advantage that they are more power-efficient and hardware friendly compared to common \glspl{ANN}. In terms of accuracy \glspl{SNN} are still behind common \glspl{ANN}, but the gap is decreasing \cite{Tavanaei.2018}. The methods used for \glspl{SNN} are very different from the methods used in conventional \glspl{DNN} and are therefore considered separately from other approaches. For this reason, \glspl{SNN} will not be discussed further here.

\subsection{Pruning}
Pruning pursues the approach of finding those weights that have little or no influence on the result and can be eliminated, therefore. That reduces network complexity and over-fitting \cite{Qiu.2016,Han.01102015}. Reducing the network complexity has two advantages: The disk storage of the network reduces and the computational effort reduces \cite{Cheng.2018, Xu.2018}. A pruned network is also called a sparse network \cite{Han.01102015}. \\
Whether pruning is permitted or not can be determined with different methods. The absolute value of the weights is found to be a proper measure \cite{guo2016dynamic,han2015learning}. Retraining is required after pruning to avoid accuracy loss \cite{guo2016dynamic,han2015learning}. An issue with retraining pruned networks can be that the gradient vanishes, which leads to slow convergence. The reason for this is simply that a large proportion of the connections is pruned. That effect can be reduced by pruning convolutional and fully connected layers separately. Additionally, slower pruning helps to increase the convergence speed during training \cite{guo2016dynamic,han2015learning}. \\
Pruning can be done in various granularity: fine-grained pruning, vector-level pruning, kernel-level pruning, group-level pruning, and filter-level pruning \cite{Cheng.2018}

\subsubsection{Fine-grained pruning}
Fine-grained pruning has no constraints. Any weight which finds to be unimportant can be pruned\cite{Cheng.2018}. Han et al. introduced a Fine-grained pruning method called Deep Compression. In the first step, the network is trained as usual. In the second step, all weights below a certain threshold are pruned. Finally, the remaining network is retrained. That method reduces, the number of parameters required by AlexNet by 9$\times$ and by VGG-16 by 13$\times$ without any drop in accuracy \cite{Han.01102015}. Incorrect pruning leads to severe accuracy loss. Therefore, caution is advised using pruning, as connections cannot be restored \cite{guo2016dynamic}. For this reason, Guo et al. introduced splicing. Splicing reconnects pruned connections, which have shown to be important. Splicing is inspired by the human brain since pruning and splicing are analogical to the synthesis of excitatory and inhibitory neurotransmitter in the human nervous system \cite{guo2016dynamic}. Based on these methods the dynamic network surgery algorithm iteratively compresses the network. Using dynamic network surgery leads to a 17.7$\times$ compression rate using AlexNet, with only a minor accuracy loss of 0.33\%.
  
\subsubsection{Vector-level and kernel-level pruning}  
In vector-level and kernel-level pruning, pruning is constrained to vectors or kernels of convolutional layer \cite{Cheng.2018}. Compared to fine-grained pruning vector-level and kernel-level pruning is more hardware friendly, since hardware accelerators are optimized for vector operations and therefore only benefit slightly from fine-grained pruning. Additionally, vector-level and kernel-level pruning require fewer indices due to de vector structure, which reduces the required storage \cite{Cheng.2018}.
  
\subsubsection{Group-level pruning}
If a filter in a convolutional layer operates on the same input feature map and provide the same pattern, the filters can be merged. Group level pruning bases on this method \cite{Cheng.2018}. Lebedev et al. introduce the group-wise brain damage algorithm for group-level pruning. It achieves a 3.2$\times$ speed up for convolutional layers of AlexNet \cite{Lebedev.2016}. 

\subsubsection{Filter-level pruning}
Filter-level pruning focus on reducing the dimension of a layer. The feature map of the following layer indicates which channel can be pruned \cite{Cheng.2018}.  

\subsection{Tensor Decomposition} \label{subsec:tensor-decomp}
As mentioned before, tensor (including matrix) computations are the basic operations required by neural networks. For this reason, tensor decomposition methods are well suited to compress and accelerate neural network models \cite{deng2020model}. Tensor decomposition has two advantages. First, it decomposes a large tensor into two smaller tensors that fit better into local memory. Second, if the original tensor doesn't have full rank, tensor decomposition reduces the computational effort, because it eliminates redundancies. 
\subsubsection{Tensor Decomposition for CNNs}
\begin{figure}[h]
	\centering
	\includesvg[width=0.6\textwidth]{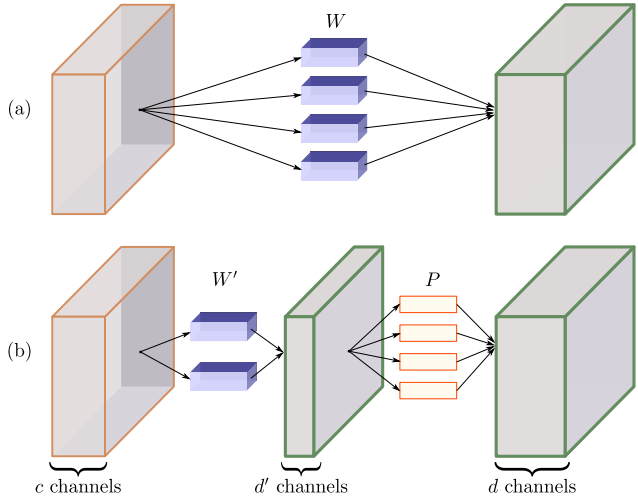}
	\caption[Decomposition of weight tensors in CNNs.]{Decomposition of weight tensors in CNNs.(a) original layer with complexity $O(dk^2c)$. (b) approximated layer width compexity reduced to $O(d'k^2c)+O(dd')$. Figure adopted from \cite[Fig. 1]{zhang2015accelerating}.}
	\label{fig:tensor-decomp}
\end{figure}

In tensor decomposition, it is assumed that the output produced by $\textbf{y}=W\textbf{x}$ can be approximately replicated by a low-rank subspace \cite{zhang2015accelerating}. This low-rank subspace is created using low-rank decomposition, which results in $y = PW'\textbf{x}+b$. That brings the benefit that the computational complexity can be reduced from $O(dk^2c)$ to $O(d'k^2c)+O(dd')$. With $c$ is the input channel size, $d$ is the original output channel size and $d'$ is the intermediate channel size \cite{zhang2015accelerating}. \Cref{fig:tensor-decomp} visualizes the principle of tensor decomposition. Zhang et al. showed that tensor decomposition can speed up the computation by a factor of $2.9$ with only a minor accuracy loss of $0.3\%$ \cite{zhang2015accelerating}. 

\subsection{Compact Models}
\begin{figure*}[tb!]
	\centering
	\includesvg[width=0.9\textwidth]{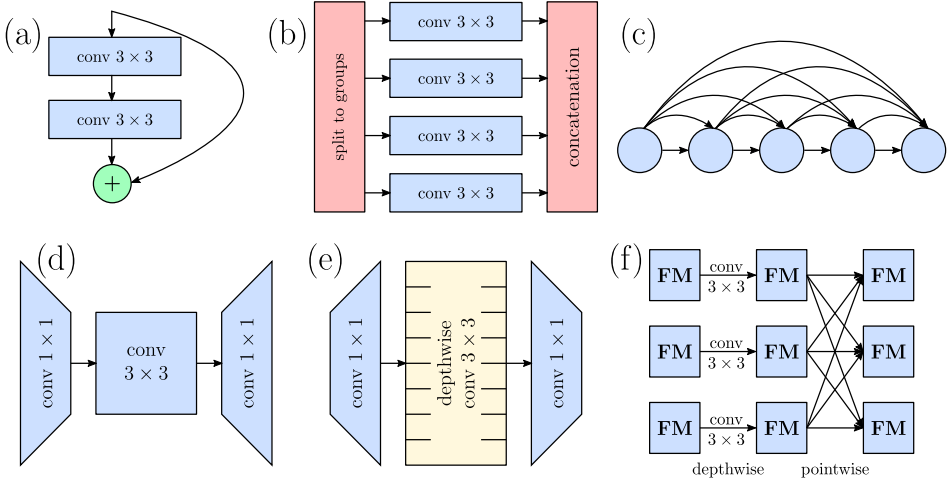}
	\caption[Compact model design elements.]{Compact model design elements.(a) Residual connection. (b) Group convolution. (c) Densely connected block.  (d) Bottleneck layer. (e) Inverse bottleneck layer. (f) Depthwise separable convolution. Adopted from \cite[Fig. 6]{deng2020model}.}
	\label{fig:sparse-elements}
\end{figure*}

In contrast to creating sparse models using pruning, compact models can be used for the targeted creation of sparse neural network models. Compact models aim to replace large kernels of convolutional layers, with smaller kernels. For example a $5\times5$ convolution can be replaced with two $3\times3$ convolutions requiring $28\%$ fewer parameter \cite{deng2020model}. However, $3\times3$ convolutions are still computationally expensive. For this reason, depthwise separable convolutions were introduced, which consists of $1\times1$ convolutions followed by depthwise separated $3\times3$ convolutions \cite{Chollet.2017}. The $1\times1$ convolutions observe the correlation of the different channels, while the $3\times3$ convolutions act as a conventional filter in the x and y-direction. This approach is similar to that of tensor decomposition in \glspl{CNN} (See \Cref{subsec:tensor-decomp}), except that this structure is specified before the training and is not generated from the trained weights. \\
Another method introduced by ResNet is to reuse the information of features generated by a former layer. That is done by adding it to the current feature map. Such networks are called residual network \cite{he2016deep}. Residual networks allow the creation of very deep networks, which yields an accuracy gain \cite{deng2020model}. \\
\Cref{fig:sparse-elements} shows how these approaches are typically realized in compact models. \Cref{fig:sparse-elements}a shows a residual connection. \Cref{fig:sparse-elements}b shows a group convolution, which splits the channels into group-wise separated channels. \Cref{fig:sparse-elements}c shows a densely connected block. \Cref{fig:sparse-elements}d shows a bottleneck layer, which downscales the channel number by using a less computational intensive $1\times1$ convolution layer. After the downscale, the computational costly $3\times3$ convolution layer performs on fewer input and output channels. After that, the channel number expands again using a $1\times1$ convolution layer. The inverse bottleneck layer visualized in \Cref{fig:sparse-elements}e, does the opposite, but it uses depthwise separated $3\times3$ convolutions. \Cref{fig:sparse-elements}f shows the difference between depthwise separated convolutions and pointwise convolutions. It shows that depthwise convolutions only consider a single input channel for computing a new output channel. In most modern compact models these approaches are combined \cite{Howard2019Mobile,redmon2018yolov3,szegedy2017inception,tan2019efficientnet}. \\
In MobileNetV2 an inverted residual block is introduced, which is a combination of an inverse bottleneck layer and a residual connection \cite{Sandler2018Mobile}. Using these blocks Tan et al. show how to find the optimum network model for given hardware resources \cite{tan2019efficientnet}. The goal of this approach is to achieve the highest possible accuracy with the lowest possible parameter number. However, the lower number of required operations of EfficientNet does not necessarily lead to a faster computation time of a single image. Bochkovskiy et al. show that the less complex network structure used by Darknet-53 leads to a lower computation time\footnote{Using an Nvidia RTX 2070 GPU in single batch mode.} compared to EfficientNet, even though nearly 5 times more computation operations are required \cite{bochkovskiy2020yolov4}.\\

\subsection{Knowledge Distillation}
The idea of knowledge distillation bases on the assumption that averaging the predictions of different models gives higher accuracy than a single model can achieve. Since that would lead to an enormous additional computational effort, knowledge distillation combines the knowledge of different networks into a new more simple network. The original models are called the teacher. The new model is called the student. Based on the output of the teacher networks, the weights and biases of the student network, are adjusted using stochastic gradient descent. Using knowledge distillation, a simple student model can achieve an accuracy that would be unachievable using direct training \cite{Sze.2017}. 

\subsection{Quantization}
Quantize data is essential in digital computations since the bit width of \glspl{ALU} is limited in all digital devices. Basically, the higher the bit width and thus the higher the computing accuracy, the greater the hardware costs \cite{Sze.2017}. Floating-point quantization allows a higher precision compared to fixed-point quantization, due to the arbitrary exponent value. However, for floating-point more complex arithmetic operations are required, compared to fixed-point operations. This lead to higher power consumption, resource costs, or latency \cite{Wang.2019}. For this reason fixed-point quantization is heavily used in \gls{ASIC}- and in \gls{FPGA}-based hardware accelerators \cite{Zhang.2016,Qiu.2016,hegde2017caffepresso,wei2017automated,guo2017angel,Nguyen.2019,Wang.2019,Chen.2016c,Chen.2017,jouppi2017datacenter,zhang2016cambricon,Han.1806201622062016}. \\
\\\
The goal in quantizing neural networks is to minimize the error between the quantized and the original network while reducing hardware costs \cite{Sze.2017}. It is possible to use distinctive types of conditioning for different layers, channels, and filter \cite{Sze.2017}. Both, the weights and the activations can be quantized. A change in the bit width of the weights has a smaller effect on the accuracy than with the activations \cite{Sze.2017}. \\
Data can be quantized linear and non-linear. Linear quantization brings the benefit that well-known calculation methods can be used, while specialized computation methods for non-linear quantized data are necessary. The advantage of non-linear quantization is that numerical ranges that occur frequently have a higher resolution than those that occur less frequently. These numerical ranges can either found by observing the distribution of weights and activations in a neural network or by learning them e.g. using k-means clustering \cite{Sze.2017}. Non-linear computations typically use \glspl{LUT}, since they offer the possibility to output a specific bit-pattern depending on an input pattern.
\\\
\subsubsection*{Binarized Quantization}
Binarized quantization is an extreme version of fixed-point quantization.
The basic idea is to constrain both weights and activations to $\{1,-1\}$ \cite{courbariaux2016binarized}. That brings the benefit that hardware-friendly XNOR operations can replace costly multiplications \cite{Liang.2018}. However, binarized quantization of both weights and activations leads to a heavy loss in accuracy \cite{Sze.2017}. For this reason, retraining is necessary for \gls{BNN} to regain some accuracy. Besides, the original is modified to increase accuracy. For example, the precision of the activations can be increased, or the first and/or the last layer are computed with full precision \cite{Sze.2017,Nguyen.2019}. \\    

Quantization of \gls{BNN} can be done using stochastic or deterministic methods. However, typically the $Sign(x)$ function is used \cite{Liang.2018,courbariaux2016binarized}: 
\begin{equation}
x_b = Sign(x) = 
\left\{  
\begin{matrix}
 +1 & x \geq 0 \\
 -1 & x < 0
\end{matrix}
\right.
\end{equation}
The problem using the $Sign(x)$ function for training, is that the derivative is a Dirac $\delta(x)$ function, which is not suitable for calculating the direction of the gradient in order to update the weight. For this reason $tanh(x)$ or $Htanh(x)$ functions are used instead of the $Sign(x)$ \cite{Liang.2018}.
\\\

\subsubsection*{Ternary Quantization}
Indifference to binarized quantization, ternary quantization adds a bit in order be able to represent a weight of 0 \cite{ZhezhiHe.2019}. That enables \glspl{TNN} to prune connections in the network. Which can increase the accuracy and significantly decrease the inference latency, since zero weights can be skipped \cite{ZhezhiHe.2019}. Originally ternary quantization uses $\{-1,0,+1\}$. However, some other constant values different from 1 are used in some related works to increase accuracy. For example, $\{-E,0,+E\}$ can be used, with $E$ is the mean absolute value of the learned weights. Another possibility is to use learned full-precision weights $\{-W^n_l,0,+W^p_l\}$ which are constant for each layer \cite{zhu2016trained}.\\
Compared to full-precision models, ternary quantization reduces the model size by a factor of 16 \cite{zhu2016trained}. Additionally if $\{-1,0,+1\}$ quantization is used, no costly multiplication operations are required \cite{zhu2016trained,ZhezhiHe.2019}. In contrast $\{-E,0,+E\}$ and $\{-W^n_l,0,+W^p_l\}$ quantization still requires multiplications. Nevertheless, constant multipliers can be used. In both cases, a more efficient hardware accelerator implementation in terms of speed, power efficiency, and area consumption is possible, with only a moderate loss in accuracy \cite{ZhezhiHe.2019,zhu2016trained}. However, compared to \glspl{BNN}, \glspl{TNN} require double the model size, but provide a gain of accuracy. Therefore, \glspl{TNN} represent a trade-off between model size an accuracy \cite{zhu2016trained}. \\
Nevertheless, retraining is required to achieve a competitive accuracy using ternary quantization, which can surpass the full-precision accuracy in some cases \cite{zhu2016trained,ZhezhiHe.2019}. \\
Zhu et al. outperformed full the precision model using ResNet-32,44,56 on CIFAR-10 dataset by 0.04\%, 0.16\%, 0.36\%, and AlexNet on ImageNet by 0.3\% using ternary $\{-W^n_l,0,+W^p_l\}$  quantization \cite{zhu2016trained}.  \\
He et al. achieved an accuracy loss of $\sim$3.9\%, 2.52\%, 2.16\% in comparison to full precision using a $\{-1,0,+1\}$ ternarized ResNet-18/34/50 on ImageNet dataset. \\
\\\
\subsubsection*{Log2 Quantization}
Log2-quantization is a subclass of non-linear quantization, which allows transforming multiplications into shifts.
The optimal quantization strategy for a single layer depends on the distribution of the weights. Typically, training uses an $L_2$ regularization, which forces gradients to favor weights close to 0. That causes the weights in a layer to be normal distributed with the mean value 0 \cite{lee2017lognet}. For this reason, an alternative for arbitrary non-linear quantization is to use logarithmic quantization, since it provides a higher resolution for small weights. In a digital system, base 2 is optimal, since it allows to transform multiplications into shifts \cite{Sze.2017,lee2017lognet}. Log2 quantization shows to achieve higher accuracy compared to linear quantization, for low-resolution weights of 4 bits and less. Additionally, area intensive multipliers change to shifts, which allows a higher performance density \cite{lee2017lognet}.


\section{Algorithmic optimization} \label{sec:algorithmic}
Algorithmic optimization applies computational transformation or vectorization of data to reduce the number of performed arithmetic operations and memory accesses. Mainly \glspl{CPU} and \glspl{GPU} use these techniques. However, various \glspl{FPGA} based neural network implementations are using algorithmic optimization \cite{Abdelouahab.26052018,mittal2018survey,Sze.2017}.

\subsection{General Matrix Multiplication}
\begin{figure*}[t!]
	\centering
	\includesvg[width=\textwidth]{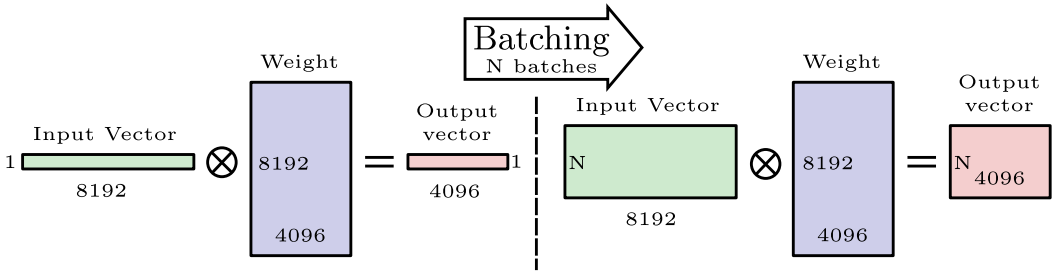}
	\caption[Batching to convert vector matrix multiplication into matrix-matrix multiplication.]{Batching to convert vector matrix multiplication into matrix-matrix multiplication. It is demonstrated that the reuse of the weight matrix increases by a factor of N if batching is used. Figure adopted from \cite[Fig. 7]{mittal2018survey}.}
	\label{fig:batching}
\end{figure*}

\Gls{GEMM} intendeds to minimize the necessary memory accesses for vector-matrix multiplications by combining multiplications with the same weight into a vector. The benefit of this method is that a single weight is fetched only once. For this reason, \gls{GEMM} implementations are most efficient for large matrices. That is the case when an entire batch of input data is computed, which is called batching. \Cref{fig:batching} visualizes that the reuse of a weight matrix increases by a factor of N for N batches. Especially for FC-layers, batching increase the utilization of \gls{SIMD} processors  \cite{mittal2018survey}. However, if real-time applications use batching, it has to be ensured that the latency requirements are not violated, since batching may increase the latency \cite{mittal2018survey}.\\
\Gls{GEMM} is widely used by \glspl{GPU}, since matrix multiplications are processed more efficiently by \gls{SIMD} and \gls{SIMT} architectures, if \gls{GEMM} is used \cite{Sze.2017,Abdelouahab.26052018,Nurvitadhi.2017}. Also for \gls{HLS} based hardware accelerator designed for \glspl{FPGA} some frameworks for using \gls{GEMM} based on \gls{OpenCL} already exists \cite{Tapiador.29092016}.

\subsection{Winograd Transformation}
\begin{figure}[h]
	\centering
	\includesvg[width=0.6\textwidth]{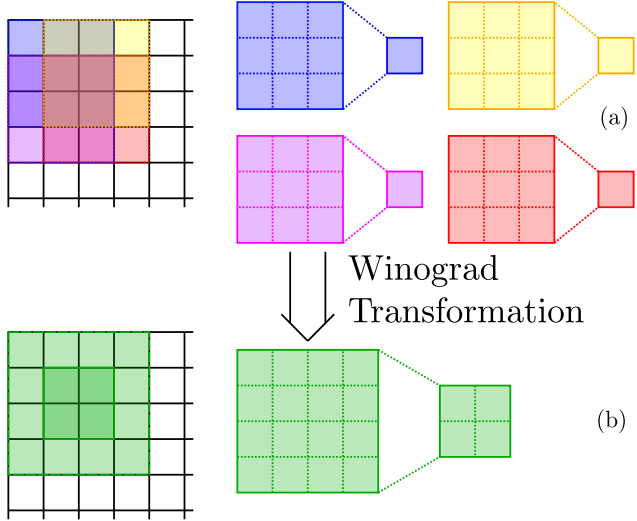}
	\caption[Winograd transformation.]{Comparison of computations required for computing 2x2 feature map tile (a) and computations required computing the same 2x2 feature map tile using Winograd-Transformation (b).}
	\label{fig:winograd}
\end{figure} 
It is a method found by Winograd. To goal is to reduce the number of multiplication required for implementing \gls{FIR} filter. Winograd transformation transforms overlapping filter kernels into non-overlapping kernels \cite{Winograd.1990}. The original method, developed for \gls{FIR} filter, can be adopted for \glspl{CNN} since the convolutional layer uses similar moving filter operations. \Cref{fig:winograd} visualizes that by transforming overlapping kernels into non-overlapping kernels, the arithmetic complexity of \Gls{CNN} is reduced by a factor up to 4 \cite{AndrewLavin.2016}.  It shows that computing a $2\times2$ output matrix requires 36 multiplications, compared to 16 multiplications needed for the same $2\times2$ output matrix using Winograd-Transformation. The efficiency increases with the kernel size since more elements are overlapping \cite{AndrewLavin.2016}.\\
\\\
The standard algorithm introduced by Winograd for a $F(2,3)$ uses $2 \cdot 3 = 6$ multiplications \cite{AndrewLavin.2016,Winograd.1990}: 
\begin{equation}
    F(2,3)= 
    \begin{bmatrix}
    d_0 & d_1 & d_2 \\
    d_1 & d_2 & d_3
   \end{bmatrix}
   \begin{bmatrix}
   g_0 \\
   g_1 \\
   g_2
   \end{bmatrix}
   =
    \begin{bmatrix}
    m_1 + m_2 + m_3 \\
    m_2- m_3 - m_4
   \end{bmatrix}   
\end{equation}
with: 
\begin{equation}
\begin{aligned}
    m_1 &= (d_0-d_2)g_0 \\
    m_2 &= (d_1 + d_2) \frac{g_0+g_1+g_2}{2} \\
    m_3 &= (d_2 - d_1) \frac{g_0-g_1+g_2}{2} \\
    m_4 &= (d_1 - d_3)g_2
\end{aligned}
\label{eq:winograd-2-3}
\end{equation}
\Cref{eq:winograd-2-3} shows that the Winograd algorithm uses only 4 multiplications, therefore it is minimal for $\mu (F(2,3))=2+3-1=4$. Additionally, 4 additions involving the data $d$, plus 3 additions, and 2 multiplications involving the filter parameter $g$, are required. However, these computations can be pre-calculated. This algorithm can also be written in matrix form \cite{AndrewLavin.2016}: 
 \begin{equation}
     Y = A^T [(G g) \odot (B^T d)]
     \label{Wino-1D}
 \end{equation}
 with $\odot$ indicating an element-wise product. The matrices for $F(2,3)$ are as follows \cite{AndrewLavin.2016}: 
\begin{equation}
\begin{aligned}
    B^T &=  
    \begin{bmatrix}
   1 & \phantom{-}0     & -1            & \phantom{-}0 \\
   0 & \phantom{-}1     & \phantom{-}1  & \phantom{-}0 \\
   0 & -1               & \phantom{-}1  & \phantom{-}0\\
   0 & \phantom{-}1     & \phantom{-}0  & -1 \\
   \end{bmatrix}
   \\
   G &= 
    \begin{bmatrix}
   1            & \phantom{-}0              & 0\\
   \frac{1}{2}  & \phantom{-}\frac{1}{2}    & \frac{1}{2} \\
   \frac{1}{2}  & -\frac{1}{2}              & \frac{1}{2}\\
   0            & \phantom{-}0              & 1 \\
   \end{bmatrix}
   \\
   A^T &= 
    \begin{bmatrix}
   1 & 1 & \phantom{-}1 & \phantom{-}0 \\
   0 & 1 & -1 & -1 \\
   \end{bmatrix}  
   \\
   g &= 
   \begin{bmatrix}
   g_0 & g_1 & g_2
   \end{bmatrix}^T
   \\
   d &= 
    \begin{bmatrix}
    d_0 & d_1 & d_2 & d_3
   \end{bmatrix}^T  
\end{aligned}
\end{equation}
\noindent
The 1D \Cref{Wino-1D} for $F(m,r)$ can be transformed into 2D for $F(m \times m,r \times r)$ as follows \cite{AndrewLavin.2016}:
 \begin{equation}
     Y = A^T [(G g G^T) \odot (B^T d B)] A
     \label{Wino-2D}
 \end{equation}
 with $g$ is a $3\times3$ filter and $d$ is a $(m+r-1) \times (m+r-1) = 4\times4$ image tile \cite{AndrewLavin.2016}:
 \begin{equation}
\begin{aligned}
   g &= 
   \begin{bmatrix}
    g_{11} & g_{12} & g_{13} \\
    g_{21} & g_{22} & g_{23} \\
    g_{31} & g_{32} & g_{33}
   \end{bmatrix}
   \\
   d &= 
    \begin{bmatrix}
    d_{11} & d_{12} & d_{13} & d_{14} \\
    d_{21} & d_{22} & d_{23} & d_{24} \\
    d_{31} & d_{32} & d_{33} & d_{34} \\
    d_{41} & d_{42} & d_{43} & d_{44} 
   \end{bmatrix}  
\end{aligned}
\end{equation}
The additions and subtractions to calculate the new $4\times 4$  weight matrix $(G g G^T)$ perform only once since they are constant for each layer. However, the new $4\times 4$ matrix $(B^T d B)$ have to be computed before the multiplications. That leads to $16 \cdot 3 = 48$ additions or subtraction required during run time for each $2\times2$ output matrix. \\

A reverse transformation is done to get the results $m$. That leads to the following additions and subtractions with $e = (G g G^T) \odot (B^T d B)$: 
\begin{equation}
    \begin{aligned}
\begin{matrix}
m_{0,0} &= e_{0, 0} + e_{0, 1} + e_{0, 2} + e_{1, 0} + e_{1, 1} + e_{1, 2} + e_{2, 0} + e_{2, 1} + e_{2, 2}\\ 
m_{0,1} &= e_{0, 1} - e_{0, 2} - e_{0, 3} + e_{1, 1} - e_{1, 2} - e_{1, 3} + e_{2, 1} - e_{2, 2} - e_{2, 3}\\
m_{1,0} &= e_{1, 0} + e_{1, 1} + e_{1, 2} - e_{2, 0} - e_{2, 1} - e_{2, 2} - e_{3, 0} - e_{3, 1} - e_{3, 2} \\

m_{1,1} &= e_{1, 1} - e_{1, 2} - e_{1, 3} - e_{2, 1} + e_{2, 2} + e_{2, 3} - e_{3, 1} + e_{3, 2} + e_{3, 3}
\end{matrix} 
    \end{aligned}
    \label{eq:wino-results-3x3}
\end{equation}
\Cref{eq:wino-results-3x3} shows that each output element requires 9 additions or subtractions.  That is equal to the original approach and doesn't influence the performance. \\
This can be summarized as follows: $F(2 \times 2, 3 \times 3)$ uses $4 \times 4 = 16$ multiplications and $16$ additions, compared to that \Cref{eq:conv2d} uses $2 \cdot 2 \cdot 3 \cdot 3 = 36$ multiplications, which reduces the amount of required multiplications by a factor of $2.25$ \cite{AndrewLavin.2016}. \\
However, the comparison is not that simple when using application-specific hardware such as \glspl{FPGA} and \glspl{ASIC},  considering that loading the data from memory is often the bottleneck. However, since less local data reuse is possible/necessary using Winograd transformation, it may not be possible to achieve a performance gain. 
Nevertheless, an advantage of the Winograd-Transformation is that, by using the local $2\times2$ output matrix, pooling can be done immediately. Since the pooling layer often follows a convolutional layer, Winograd-Transformation can reduce the required data movement. \\
Winograd transformation is used in \glspl{GPU}, \glspl{FPGA} and \glspl{ASIC}. It increases throughput as well as power efficiency. 
For the use of Winograd transformation in \gls{FPGA} and \gls{ASIC} implementations, it is recommended to consider it already in the hardware design phase,  to optimally design the data flow and the processing elements.  

\subsection{Fast Fourier Transformation}
\begin{figure}[h]
	\centering
	\includesvg[width=0.6\textwidth]{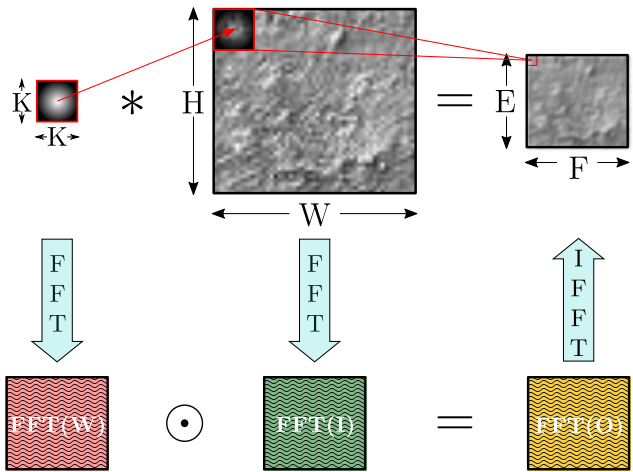}
	\caption[Fast Fourier Transformation to accelerate CNN.]{Fast Fourier Transformation used to accelerate CNN. By applying FFT to the filter and to the input image the convolution operation can be transformed into an element-wise multiplication. Figure adopted from \cite[Fig. 20]{Sze.2017}.}
	\label{fig:fft}
\end{figure} 

\Gls{FFT} is an algorithm to transform signals from image to time domain and vice versa. Convolutions in the time domain correspond to multiplications in the image domain and vice versa. \Cref{fig:fft} visualizes the transformation of a convolution operation in a convolutional layer into an element-wise multiplication using \gls{FFT} \cite{Sze.2017,Abdelouahab.26052018}.  A 2D \gls{FFT} is computed by calculating first a 1D \gls{FFT} of every row followed by a 1D \gls{FFT} of every column \cite{mittal2018survey}. \Gls{FFT} reduces the arithmetic complexity from $O(N_K^2 N_O^2) \rightarrow O(N_O^2 log_2 (N_O))$ with $N_K \times N_K$ is the kernel size and $N_O \times N_O$ is the output size \cite{Sze.2017,Abdelouahab.26052018}. The benefits of \gls{FFT} decreases with the kernel size \cite{Sze.2017}. It is also difficult to combine \gls{FFT} with sparsity, which often offers higher profits \cite{Sze.2017}. \\
Since modern \glspl{CNN} mainly use small kernel sizes \gls{FFT} has lost importance in modern hardware accelerators.

\section{Accelerating nested loops} \label{sec:LoopOptimization}
The idea of neural network hardware accelerators is to parallelize operations. From an algorithmic point of view, matrix computations can be seen as nested loops. There are 3 approaches to accelerate the computation of these loops: Loop recording, unrolling, and pipelining. Loop tiling, on the other hand, deals with efficient memory allocation  \cite{mittal2018survey}. 
\subsection{Loop recording}
It is also known as loop interchange \cite{Cheng.2018}. The goal is to rearrange the processing of nested loops, to avoid redundant memory accesses, and to maximize cache usage efficiency \cite{mittal2018survey}. That can lead to a significant performance gain \cite{Cheng.2018}. 
\subsection{Loop unrolling}
Independent loops can execute simultaneously using parallel hardware. The goal is to decrease the computation time by increasing the hardware utilization \cite{Cheng.2018,mittal2018survey}. Loop unrolling defines the optimal \gls{PE} structure \cite{Ma.2018}. 
\subsection{Loop pipelining}
Pipelining divides nested loops into sequential steps, which can be performed in parallel \cite{mittal2018survey}. The possibility of pipelining depends on the architecture of the hardware accelerator. 
\subsection{Loop tiling}
Loop tiling partitions the cache into tiles, which load as a whole chunk from memory, to avoid that data in the cache to supersede before the usage \cite{mittal2018survey}. In terms of \gls{ASIC}- and \gls{FPGA} based hardware accelerators, loop tiling deals with the efficient usage of available on-chip memory to increase the locality of data \cite{Ma.2018}.

\section{Graphics Processing Unit} \label{sec:gpu} 
\Glspl{GPU} enabled the great breakthrough of \glspl{DNN} in the field of computer vision, since \glspl{GPU} provide a massive parallelization of the computations required by \glspl{DNN} such as AlexNet \cite{AlexKrizhevsky.2012}. For this reason \Glspl{GPU} are the most widely used hardware accelerator used by \glspl{DNN} architects \cite{Nurvitadhi.2017,Sze.2017,Abdelouahab.26052018}. \\
Compared to \glspl{ASIC} and \glspl{FPGA}, \glspl{GPU} provide a high throughput and are easy to use. However, recent innovations in the structure of neural networks, such as sparsity, \glspl{BNN} and \glspl{SNN} result in a week algorithmic efficiency using \glspl{GPU}. This structures can be handled more efficiently using \glspl{FPGA} and \glspl{ASIC} \cite{Xu.2018,Nurvitadhi.2017}.

\subsection{Frameworks}
\Glspl{GPU} are very popular since many frameworks already exist which offer a high-level API for creating, training, and testing neural networks. Some examples are: Caffe, Torch, Theano, Caffe2, PyTorch, TensorFlow, MXNet, CoreML, CNTK, and TensorRT \cite{venieris2018toolflows}.
This simplification and the support of training algorithms due to the floating-point computation is one of the reasons why most \glspl{DNN}, which have won the ImageNet competition, use \glspl{GPU}
\cite{Xu.2018,Abdelouahab.26052018,szegedy2017inception,szegedy2015going,AlexKrizhevsky.2012,Simonyan.04092014,he2016deep}.

\subsection{Architecture}
\begin{figure}[h]
	\centering
	\includesvg[width=0.38\textwidth]{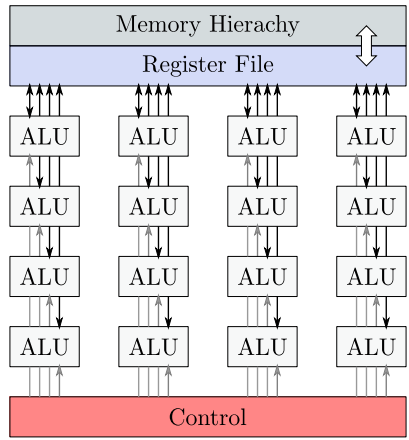}
	\caption[Temporal \gls{SIMD}/\gls{SIMT} architecture]{Temporal \gls{SIMD}/\gls{SIMT} architecture. Adapted from \cite[Fig. 17]{Sze.2017}.}
	\label{fig:simd}
\end{figure}

\glsreset{GEMM}
\glspl{GPU} consists of many floating-point arithmetic units for vector processing in combination with a high bandwidth memory. This is essential for \glspl{DNN}, since most \glspl{DNN} are based on dense floating-point \glspl{GEMM} of 32-bit floating-point data \cite{Sze.2017,Nurvitadhi.2017}.   
\Cref{fig:simd} visualizes a typical temporal \gls{SIMD}/\gls{SIMT} structure used by \glspl{GPU}. Temporal \gls{SIMD} architectures utilize a centralized control for a large number of \glspl{ALU} which share registers and memory. In temporal architectures, each \gls{ALU} can fetch data from memory and load data back to memory. However, a single \glspl{ALU} in a \gls{GPU} cannot communicate with each other directly. For this reason, communication between individual \glspl{ALU} is done using shared memory. This bottleneck increases the memory traffic, which is power intensive and decreases the possible throughput \cite{Sze.2017}. Compared to that spatial structures implemented in \glspl{FPGA} and \glspl{ASIC} avoid this bottleneck \cite{Sze.2017}.

\subsection{Embedded GPUs}
Embedded \gls{GPU} based hardware accelerator provides an easy to use, cost-efficient platform for fast development since many popular frameworks for neural networks are also suitable for embedded \glspl{GPU}. Platforms like NVIDIA Jetson combine an ARM \gls{CPU} with a small power-optimized \gls{GPU}, for parallel image processing \cite{hegde2017caffepresso}. 

\subsection{Overview}
\Cref{tab:GPU-overview} presents an overview of existing Nvidia \glspl{GPU}. It shows that the introduction of tensor processing units, specially designed for \gls{AI}, leads to huge performance gains. Besides, smaller data types are supported, which offers an additional performance gain. 

\begin{table*}
	\centering
	\caption{Overview of GPU-based hardware accelerator implementations.}	
	\small
	\setlength{\extrarowheight}{.5em}
	\begin{tabularx}{\textwidth}{c c c c c c c c c c }
		\toprule[1.5pt]
		\multirow{2}{*}{Name} & Area & feature & Quanti- & Bit & Tensor & Throughput & Freq. & Power & \multirow{2}{*}{$[\frac{\si{GOPS}}{\si{\mm\squared}}]$\textsuperscript{(b)}} \\
		& [\si{\mm\squared}] & size [\si{nm}] & zation& width & unit & [\si{TOPS}]\textsuperscript{(a)}& [\si{MHz}] & [\si{W}] & 
		\\
		\midrule[1.5pt] 
		\multirow{3}{*}{V100$^1$ \cite{NVIDIA.2017}} & \multirow{3}{*}{815} & \multirow{3}{*}{12} & float & 64 & & 7.8 & 1530 & 300 & 9.57 \\
		 & & & float & 32 & & 15.7 & 1530 & 300 & 19.26 \\
		 & & & mixed & 32-8 & X & 125 & 1530 & 300 & 153.37 \\ 
		\midrule
		\multirow{4}{*}{T4$^1$ \cite{NVIDIA.turing}} & \multirow{4}{*}{545} & \multirow{4}{*}{12} & float & 32 &  & 8.1 & 1590 & 70 & 14.81 \\
		& & & float & 16 & X & 65 & 1590 & 70 & 119.26 \\
		& & & fixed & 8 & X & 130 & 1590 & 70 & 238.53 \\
		& & & fixed & 4 & X & 260 & 1590 & 70 & 477.06 \\
		\midrule
		 & \multirow{4}{*}{545} & \multirow{4}{*}{12} & float & 32 &  & 10.6 & 1710 & 225 & 19.45 \\
		RTX 2080$^2$ & & & float & 16 & X & 84.8 & 1710 & 225 & 155.6 \\
		\cite{NVIDIA.turing} & & & fixed & 8 & X & 169.6 & 1710 & 225 & 311.19 \\
		& & & fixed & 4 & X & 322.2 & 1710 & 225 & 591.2 \\
		\midrule
		 & \multirow{4}{*}{628.4} & \multirow{4}{*}{8} & float & 32 &  & 29.8 & 1710 & 320 & 47.42 \\
		RTX 3080$^2$ & & & float & 16 & X & 119 & 1710 & 320 & 189.3 \\
		\cite{nvidia.ampere} & & & fixed & 8 & X & 238 & 1710 & 320 & 378.7 \\
		& & & fixed & 4 & X & 476 & 1710 & 320 & 757.47 \\
		\midrule		
		\multirow{2}{*}{TX 1$^3$ \cite{NVIDIA.2015}} & \multirow{2}{*}{N/A} & \multirow{2}{*}{20} & float & 32 &  & 0.512 & 998 & 15 & N/A \\
			& & & float & 16 & & 1.024 & 998 & 15 & N/A \\
		\midrule
		AGX Xavier$^4$ &   \multirow{2}{*}{N/A} & \multirow{2}{*}{12} & float & 16 & X & 11 & 1370 & 30 & N/A \\
		\cite{JetsonAGX.nvidia}& & & int & 8 & X & 22 & 1370 & 30 & N/A \\		
		\midrule
		\midrule
		\multicolumn{10}{l}{(a) The fictitious peak performance is used due to a lack of comparable data} \\
		\multicolumn{10}{l}{(b) Performance density} \\
		\multicolumn{10}{l}{NVIDIA $^1$Tesla, $^2$GeForce, $^3$Jetson (Tegra), $^4$Jetson (Volta)}  \\
		\bottomrule[1.5pt]
	\end{tabularx}
	\label{tab:GPU-overview}
\end{table*}
\section{Application Specific Integrated Circuits} \label{sec:asic}
\Glspl{ASIC} provide the greatest degree of freedom for designing a neural network hardware accelerator. \gls{ASIC} designs require a huge effort and need an in-depth knowledge of chip design and neural networks \cite{mittal2018survey,Wang.2019}. However, due to the great design freedom of \glspl{ASIC}, \gls{ASIC} based hardware accelerators can outperform \glspl{GPU} and \glspl{FPGA} in both speed, and power efficiency \cite{Xu.2018,Wang.2019}. The big disadvantage of \glspl{ASIC} is a long time to market, and the high initial costs, which hinder them from keeping up with the rapid progress in the field of deep neural networks \cite{Wang.2019}.
\\
\Cref{tab:ASIC-overview} gives an overview of existing \gls{ASIC} based hardware accelerators. It shows that DaDianNao achieves the highest performance density and has moderate power consumption. However, the metrics are given cryptically as a multiple of the GPU performance. 

\begin{table*}
	\centering
	\caption{Overview of existing ASIC-based hardware accelerator implementations.}	
	\small
	\setlength{\extrarowheight}{.5em}
	\begin{tabularx}{\textwidth}{c c c c c c c c c }
		\toprule[1.5pt]
		\multirow{2}{*}{Name} & Area & feature & Quanti- & Bit  & Throughput & Frequency & Power & \multirow{2}{*}{$[\frac{\si{GOPS}}{\si{\mm\squared}}]$\textsuperscript{(b)}} \\
		& [\si{\mm\squared}] & size [\si{nm}] & zation & width & [\si{\gls{GOPS}}]\textsuperscript{(a)}& [\si{MHz}]  & [\si{mW}] & 
		\\
		\midrule[1.5pt] 
		\footnotesize{DaDianNao \cite{chen2014dadiannao}} & 4335* & 28 & fixed  & 16 & 1586288* & 606 & 48380* & 366* \\
		\midrule
		\footnotesize{EIE \cite{Han.1806201622062016}} & 40.8 & 45 & fixed & 16 & 102 & 800 & 590 & 2.5 \\
		\midrule
		\footnotesize{Cambricon-X \cite{zhang2016cambricon}} & 6.38 & 65 & fixed & 16 & 544 & 1000 & 954 & 85.26 \\
		\midrule
		\footnotesize{Eyeriss \cite{Chen.2017}} & 16 & 65 & fixed & 16 & 67.2** & 200 & 278 & 4.2\\
		\midrule
		\footnotesize{TPU \cite{jouppi2017datacenter}} & 331*** & 28 & fixed  & 8 & 92000 & 700 & 40000 & 69.48 \\ 
		\midrule
		\midrule
		\multicolumn{9}{l}{(a) The fictitious peak performance is used due to a lack of comparable data} \\
		\multicolumn{9}{l}{(b) Performance density} \\
		\multicolumn{9}{l}{*For 64 nodes. In paper only stated as a multiple of NVIDIA K20M} \\
		\multicolumn{9}{l}{**Stated in the paper in GMAC. Therefore the double value is used} \\
		\multicolumn{9}{l}{***4 dies} \\
		\bottomrule[1.5pt]
	\end{tabularx}
	\label{tab:ASIC-overview}
\end{table*}

\subsection{DianNao family}
The architecture of DianNao uses global input data, weight, and output data buffers in combination with a \gls{NFU} for computations. The \gls{NFU} consists of 16-bit fixed-point multipliers in the first stage, an adder tree in the second stage, followed by a non-linearity unit. The second stage additionally contains shifters and max operators for pooling layers. The goal of DianNao is to support neural networks of any scale \cite{Chen.2016c,Chen.2014}. 
In DaDianNao, adding additional buffers next to the neurons reduces the data movement. Additionally, the single computational block is split into four identical \glspl{NFU}. DaDianNao targets high performance \cite{Chen.2016c,chen2014dadiannao}. ShiDianNao improves the DianNao structure by using the fact that in \glspl{CNN} a whole vector shares the same weight. This condition is used to reduce the required DRAM accesses and to reduce the power consumption, therefore \cite{Chen.2016c}. 

\subsection{EIE: Efficient Inference Engine}
EIE is designed for operating on sparse networks. That brings the benefit that faster and more power-efficient SRAM replaces power-hungry DRAM memory. The architecture of EIE utilizes a central control unit, which controls a \gls{PE} array. Each \gls{PE} computes one slice of the compressed network. The central control unit broadcasts non-zero activation values to the \glspl{PE}, which use an activation queue, for buffering input data \cite{Han.1806201622062016}.  
\subsection{Eyeriss}
The core architecture of Eyersiss is a spatial array with 168 \glspl{PE} in a 12$\times$14 shape, including a 108-kB GLB, an RLC CODEC, and a ReLU module. Each \gls{PE} can communicate with the neighboring \glspl{PE} or the GLB. A \gls{PE} contains a \gls{MAC} unit a local buffer called spads and a control unit \cite{Chen.2017}.  
\subsection{Tensor processing unit}
A matrix multiply unit, which includes 256$\times$256 \glspl{MAC}, is the core element of the tensor processing unit. The multiplication unit has an input bit width of 8, the accumulation unit a bit width of 16. The matrix multiplication unit is followed by 4096, 256-element, 32-bit accumulators. Additionally, a dedicated activation module for applying the non-linearity and a dedicated normalization and pooling module is used. A 24 MiB on-chip buffer is used for storing the results. The weights are loaded directly from DRAM into a specialized weight FIFO \cite{jouppi2017datacenter}.  

\section{Field Programmable Gate Arrays} \label{sec:fpga}
\Glspl{FPGA} offer a cost-efficient method to develop custom hardware solutions since \glspl{FPGA} provide the possibility to implement a reprogrammable logic circuit. Realizing a task in a logic circuit offers the advantage of excessive parallelization compared to a software solution in a \gls{CPU}, which leads to a considerable speed advantage and higher power efficiency \cite{wang2018survey,Guo.24122017}. Additionally, the implementation of an accelerator in hardware reduces the necessary memory accesses since the intermediate results can be passed on directly to the next \gls{PE}. In addition to \glspl{CLB}, modern \glspl{FPGA} implement hundreds to thousands of \glspl{DSP} and various options for storing data locally. Embedded \glspl{BRAM} and FIFOs can be used for larger amounts of data. In contrast, \glspl{LUT} are well suited to be used as distributed RAM for smaller amounts of data. That enables \gls{FPGA} based neural network implementations to minimizes the delay due to memory access. \\
For this reason \glspl{FPGA} offer the possibility to realize a low latency inference in combination with a high energy efficiency $(\sim 10-50 \si{GOP/s/W})$ \cite{Ma.2018}. All in all, \glspl{FPGA} represent a tunable balance between performance and energy consumption \cite{venieris2018toolflows}. However, greater freedom of design also inevitably leads to a more complex solution and more development effort. 

\subsection{Algorithmic optimization in FPGA based hardware accelerators}
\gls{FPGA} based hardware accelerators widely use the algorithmic optimization methods described in \Cref{sec:algorithmic}. However, the efficiency of these methods depends on the structure of the hardware accelerator. \Gls{GEMM} is well suited for architectures that support batching, such as single computation engines. Nevertheless, for streaming architectures, \gls{GEMM} is not applicable since multiple batches would require an enormous tile size of the buffers \cite{Ma.2018}. \\
Basically \glspl{FPGA} would be well suited for the use of \gls{FFT}. However, \gls{FFT} is more efficient for large kernel sizes and the trend in \gls{DNN} structures is towards small kernel sizes \cite{Sze.2017,he2016deep,szegedy2017inception,redmon2018yolov3}. 

\subsection{Mapping Convolutional Neural Networks on FPGAs} \label{subsec:MappingFPGA}
The task of mapping a \gls{DNN} having millions of parameter and ten to hundred layers is a complex multidimensional optimization problem \cite{Ma.2018}. The goal is to minimize off-chip memory access since these lead to high energy consumption and latency \cite{Ma.2018,Sze.2017}. However, high hardware utilization is required to achieve high throughput. That is achieved by using loop optimization methods such as loop unrolling, loop tiling, loop recording, and loop pipelining (See \Cref{sec:LoopOptimization}). \\
On-chip memory is usually not sufficient for the large amounts of data required for input data, temporary results, and weights of a \glspl{DNN}. For this reason, the following three typically memory hierarchies are used: External memory, on-chip buffers (cache), and registers for the individual \glspl{PE}. \\
Since full unrolling of large \glspl{DNN} is often not possible due to the limited resources, intermediate results are buffered locally. The less intermediate results are buffered, and the earlier they can be processed, the less data movement is required. That leads to fewer off-chip memory accesses, and the efficiency increases, therefore. Additionally, suitable tile size is necessary to store the intermediate results \cite{Ma.2018}. \\
\\\
\glsreset{HLS}
A difficulty for the use of \glspl{FPGA} is the complex design flow to get an efficient implementation of a neural network structure in an \gls{FPGA}. \Gls{HLS} based on widely used programming languages such as C, C++, and OpenCL promises to simplify and accelerate the design flow. These offer the possibility of automating the design flow based on the parameters of each layer. That enables neural network designers without any hardware experience to create a customized hardware implementation \cite{venieris2018toolflows}. Venieris et al. provide an overview of existing CNN-to-FPGA tool flows, and features such as performance, arithmetic precision, portability supported NN-models, and optimization is compared \cite{venieris2018toolflows}. However, abstraction comes with a loss in efficiency \cite{Nguyen.2019}.

\subsection{Hardware architectures}
Typical automated toolflows use one of following 2 hardware structures: Single computation engines and streaming architectures \cite{venieris2018toolflows}.

\begin{figure}[h]
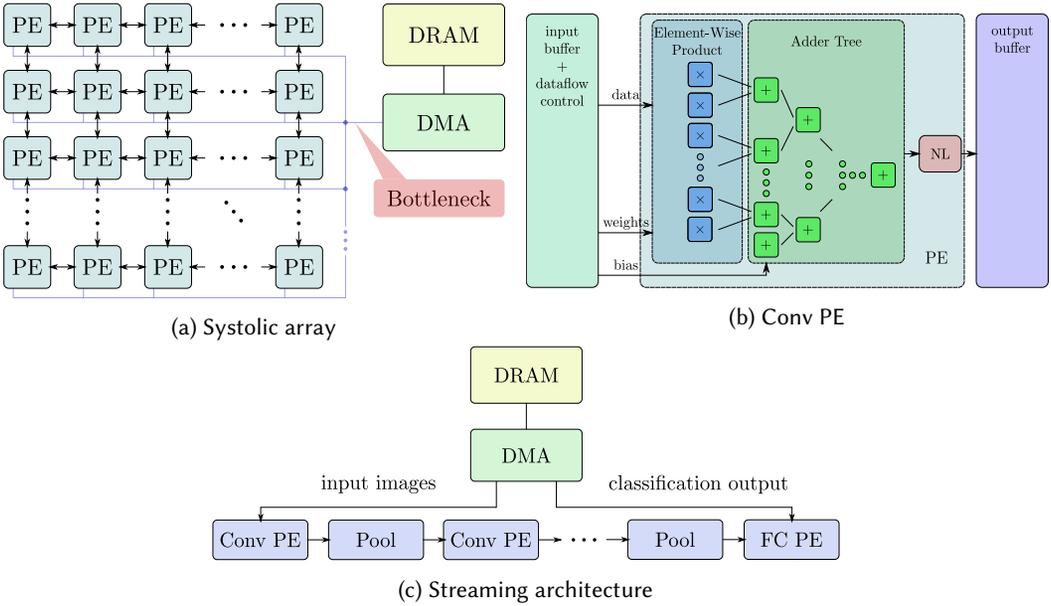

	\centering
	\begin{subfigure}[a]{0.48\textwidth}
		\includesvg[width=\textwidth]{graphics/systolic-array.svg}
		\caption{Systolic array}
		\label{subfig:systolic}
	\end{subfigure}%
	\hfill
	\begin{subfigure}[a]{0.50\textwidth}
		\includesvg[width=\textwidth]{graphics/conv-pe.svg}
		\caption{Conv PE}
		\label{subfig:conv-pe}
	\end{subfigure}%
	\hfill
	\begin{subfigure}[a]{0.6\textwidth}
		\includesvg[width=\textwidth]{graphics/streaming-arch.svg}
		\caption{Streaming architecture}
		\label{subfig:streaming}
	\end{subfigure}%
	\caption[Typical hardware architectures used in an FPGA.]{Typical hardware architectures used in an \gls{FPGA}-based hardware accelerator. \subref{subfig:systolic} Typical structure of systolic array. \acrshortpl{PE} can communicate with adjacent \acrshortpl{PE}. Weights, input, and output data are transferred using \acrshort{DMA}. Adapted from \cite[Fig. 4]{Abdelouahab.26052018}. \subref{subfig:streaming} Typical structure of a streaming architecture. Each \gls{PE} computes an independent layer. The data is pipelined throw the accelerator. Adapted from \cite[Fig. 1]{venieris2018toolflows}. \subref{subfig:conv-pe} Typical structure of convolutional processing element. It consists of $N$ multipliers depending on the kernel size, followed by an adder tree, to sum, the intermediate results and a non-linearity element. Adapted from \cite[Fig. 4]{Qiu.2016}.} 
	\label{fig:systolic-arr}
\end{figure} 

\subsubsection{Single computation engines}
Single computation engines are usually implemented in a systolic array \cite{venieris2018toolflows}. \Cref{fig:systolic-arr} visualizes the typical structure of systolic arrays. It shows that in a systolic array, each \gls{PE} can exchange data with the adjacent \glspl{PE}. Which enables systolic arrays to achieve a high data rate between the \glspl{PE}. The \glspl{PE} of systolic arrays can either consist of \gls{MAC}-units, convolutional \glspl{PE} which is shown in \Cref{subfig:conv-pe} or of \gls{SIMD} units \cite{venieris2018toolflows,wei2017automated}. However, weights, inputs, and outputs are transferred from or to the external memory, which leads to a bottleneck. Adding additional buffers for weights and input data can reduce problem\cite{wei2017automated}. The benefit of using systolic arrays is that they can adapt easily for different \gls{DNN} structures. However, the flexibility comes with the price of reduced efficiency, which leads to a very different throughput using different \glspl{DNN} \cite{Venieris.2019,guo2017angel}.

\subsubsection{Streaming architectures} 
\Cref{subfig:streaming} visualizes the structure of streaming architectures, which implement an independent specialized \gls{PE} for each layer. The data is pipelined throw the hardware accelerator. However, It has to be ensured that the \glspl{PE} are designed so that the calculations of individual layers take the same amount of time to prevent idling. The weights are stationary required for a specific layer, which reduces global data transfer. That enables streaming architectures to implement a highly efficient accelerator. However, compared to single computation engines, streaming architectures are less flexible since they are optimized for a single \gls{DNN} structure. Additionally, it is difficult to increase or decrease the size of the accelerator, if the \gls{FPGA} type changes \cite{venieris2018toolflows}.

\subsubsection*{Processing elements} 
\glspl{MAC} are heavily used in neural networks since it is the basic operation for fully connected and convolutional layer. As the name indicates \glspl{MAC} consists of a multiplier and an accumulator \cite{Sze.2017}. However, when accumulating the results of different input channels of a \gls{CNN}, an accumulation without a multiplication is required. In this case, costly multipliers are unused. For this reason, especially in streaming architectures, a specialized convolutional \gls{PE} is used, which consists of $N$ multipliers depending on the kernel size, followed by an adder tree, to sum the intermediate results of each input channel. Usually, a non-linearity unit follows the adder tree \cite{Ma.2018,Qiu.2016}. Additionally, the latency decreases by using Conv-\glspl{PE} since less pipelined stages are required. \Cref{subfig:conv-pe} visualizes the structure of a convolutional processing element \cite{Qiu.2016}.

\subsection{Overview of existing implementations}
In 2015 Zhang et al. introduced a 32-bit floating-point single computation engine on a high-end Xilinx \gls{FPGA} using \gls{HLS}. The accelerator uses 64 two-level unrolled \glspl{PE} operating with 100\si{MHz}. Each \gls{PE} has 7 inputs for activations and weights and 1 bias input. Additionally, buffers are applied to input and output. The size of on-chip buffers is determined using loop tiling. A MicroBlaze \gls{CPU} controls the data flow of the hardware accelerator. The system has a power consumption of 18.61\si{W}. Due to the high precision \glspl{PE}, no accuracy loss occurs. However, the high precision results in a low throughput of 61.62\si{\gls{GOPS}} \cite{Zhang.2015}. For this reason, this design has compared to others a poor implementations efficiency. \\
In 2016 Zhang et al. created a hardware-software co-design library called Caffeine, which reduces the design effort for creating a hardware accelerator based on Caffe. It has similar performance metrics compared to the previous version.  However, in this version 16 bit fixed point quantization is available, which increases the performance by a factor of 3.8. \\ 
\\\
Also, in 2016 Qiu et al. presented a 16-bit fixed-point single computation engine specialized for convolution operations. This structure brings the benefit that convolutions computations are more efficient. However, due to the non-optimal data flow for FC layers, they can only be calculated with low-performance \cite{Qiu.2016}. The overall throughput is similar to the 16-bit fixed-point implementation introduced by Zhang et al. \cite{Qiu.2016,Zhang.2016}. Additionally, the throughput of a 4/8 bit Quantization is estimated, which would achieve a 2.6 $\times$ higher throughput with a reasonable low accuracy loss of 0.4\% \cite{Qiu.2016}. \\
\\\
Wei et al. created an automated systolic array architecture synthesis in 2017. The unique thing in this implementation is that each \gls{PE} contains a \gls{SIMD} vector accumulator. The systolic array structure aims to maximize the on-chip data reuse. For this reason, additional buffers for input, output, and weights are placed at the systolic array edges \cite{wei2017automated}. This implementation achieves compared to Zhang et al. a very high throughput of 461\si{\gls{GOPS}} using 32-bit floating-point, and 16-bit fixed-point implementation \cite{wei2017automated,Zhang.2016,Zhang.2015}. However, a comparison of the implementations is difficult since different \gls{FPGA} vendors are used which utilize different \gls{DSP} structures. \\
\\\
Aydonat et al. use 16-bit floating-point operations for accumulation and 16 bit fixed point operation for multiplication to reduce the overhead from using FP32 \glspl{DSP}. The hardware accelerator structure uses serial \glspl{PE}, optimized for computing Winograd transformed feature maps \cite{Aydonat.2017}. This design outperforms the 16-bit implementation of Wei et al. by a factor of 1.18 in terms of throughput \cite{Aydonat.2017,wei2017automated}. \\
\\\
Liang et al. introduced in 2018 a \gls{BNN} for FPGA. Bit-level XNOR and shift operations replace costly multiplications, due to the binarization of both weights and activation. That enables a massive boost in throughput. However, an extensive drop in accuracy occurs for complex \glspl{DNN} such as AlexNet. The use of full precision computations in the last layer counters this effect. Nevertheless, an accuracy loss of 14.7\% occurs using AlexNet \cite{Liang.2018}. \\
\\\
Chung et al. and Fowers et al. presented the Brainwave neural processing unit (NPU) in 2018. Which is a single thread \gls{SIMD} instruction set architecture paired with 96000 \glspl{MAC} for vector-matrix and vector-vector operations. The target of the hardware accelerator is a maximal performance for cloud-scale real-time applications. The quantization is a 9-bit pseudo floating point quantization, including a sign bit, a 5-bit exponent, and a 2-5 bit mantissa. While 128 numbers share a common exponent. This quantization results in an accuracy loss of 1-2\%. When performing ResNet-50 on an Arria 10 1150, the brainwave NPU achieves a lower real-time latency (batch 1) than an Nvidia P40 GPU using int8 quantization \cite{fowers2018configurable,chung2018serving}.\\
\\\
Nguyen et al. present a real-time YOLO object detection streaming architecture on a Xilinx XC7V485. A specialized \gls{PE} is implemented for each layer. The images are pipelined throw the processing elements. A binary quantization of the weights and a 4 to 6-bit quantization of the activations are used to minimize the footprint of the accelerator. That leads to an accuracy loss of 2.5\%. This implementation achieves a 1.24$\times$ higher frame rate with an 11.5$\times$ higher energy efficiency compared to an NVIDIA GTX Titan X \gls{GPU} using Sim-YOLO-v2 \cite{Nguyen.2019}.  \\
\\\
\Cref{tab:FPGA-overview} provides an overview of the introduced related \gls{FPGA} based hardware accelerator implementations. It compares the design methodology and the quantization with their influence on the throughput and utilization efficiency to estimate implementation efficiency. The throughput divided by the available \glspl{LE} is used as a measure for the utilization efficiency because it shows how efficiently the available resources are used to generate maximum throughput. In some related works, the throughput divided by the number of \glspl{DSP} is used for implementation efficiency \cite{venieris2018toolflows}. However, the number of \glspl{DSP} and \glspl{LE} correlate anyway, which leads to a similar comparison. Historically, a single \gls{LE} is defined as a four-input \gls{LUT}, a programmable register, and a carry chain. However, in modern \glspl{FPGA} \glspl{LUT} usually have more than 4 inputs. For this reason, an adjustment factor is used for the conversion into a fictitious number of 4 input \glspl{LUT}. Nevertheless, a comparison of implementations using different \gls{FPGA}-vendors is difficult anyway. \\
\\\
From \Cref{tab:FPGA-overview} it can be observed that the influence of using a \gls{HLS} or \gls{RTL} based design methodology has a minor impact on the implementation efficiency. Furthermore, quantization is crucial for implementation efficiency. However, the efficiency gain comes with an accuracy loss. The throughput gain is much higher than the accuracy loss if a suitable combination of weight and activation bit width is used, though. Furthermore, latest implementations show that \glspl{FPGA} are able to challenge \glspl{GPU} in real-time computation tasks \cite{Nguyen.2019,fowers2018configurable}. 
\begin{table*}
	\centering
	\caption{Overview of existing FPGA-based hardware accelerator implementations.}

	\footnotesize
	\setlength{\extrarowheight}{.5em}
	\begin{tabularx}{\textwidth}{c c c c c c c c c c c}
		\toprule[1.5pt]
		\multirow{2}{*}{Article}  & Quant- & \multicolumn{2}{c}{Bit width}  & Accuracy & Throughput & Freq. & Power & \multicolumn{2}{c}{FPGA} & \multirow{2}{*}{$\frac{\mbox{GOPS}}{\mbox{k LE}}$} \\
		 & ization& W &	X & loss\textsuperscript{(a)} & [\si{\gls{GOPS}}]& [\si{MHz}] & [\si{W}] & Type & \# \gls{LE} & \\

		\midrule[1.5pt] 
		\scriptsize{Zhang et al. \cite{Zhang.2015}}  & float & 32 & 32 & 0\% & 61.62 & 100 & 18.61 & \scriptsize{XC7VX485} & 485.7k & 0.13 
		\\
		\midrule[0.1pt]
		\multirow{2}{*}{\scriptsize{Zhang et al. \cite{Zhang.2016}}} & float & 32 & 32 & 0\% & 96 & \multirow{2}{*}{200} & 25 &\multirow{2}{*}{\scriptsize{XCKU060}} & \multirow{2}{*}{726\si{k}} & 0.13 \\
		&  fixed & 16 & 16 & N/A & 365 & & 25 &  & & 0.5
		\\
		\midrule[0.1pt]
		\multirow{2}{*}{\scriptsize{Qiu et al. \cite{Qiu.2016}}} & \multirow{2}{*}{fixed} & 16 & 16 & 0\% & 187.8 & \multirow{2}{*}{150} & 9.63 &  \multirow{2}{*}{ \scriptsize{XC7Z045}} & \multirow{2}{*}{350\si{k}} & 0.54 \\
		&  & 4 & 8 & 0.4\% & 495** & & N/A  &  &  & 1.41
		\\
		\midrule[0.1pt]
		\scriptsize{Hegde et al. \cite{hegde2017caffepresso}} & fixed & 16 & 16 & N/A & 11.5 & 180 & 19 & \scriptsize{XC7Z045} & 350\si{k} & 0.03
		\\ 
		\midrule[0.1pt]
		\multirow{2}{*}{\scriptsize{Wei et al. \cite{wei2017automated}}} & float & 32 & 32 & 0\% & 461 & 221.65 & N/A & \multirow{2}{*}{\scriptsize{Arria 10}} & \multirow{2}{*}{1150\si{k}} & 0.4 \\
		& fixed & 8 & 16 & 2\% & 1171& 231.85 & N/A & & & 1.02
		\\  
		\midrule[0.1pt]
		\multirow{2}{*}{\scriptsize{Guo et al. \cite{guo2017angel}}} & \multirow{2}{*}{fixed} & 16 & 16 & 0.06\% & 187.80 & 150 & 9.63 & \scriptsize{XC7Z045} & 350k  & 0.54 \\
		& & 8 & 8 & 0.62\% & 19.2 & 100 & 2 & \scriptsize{XC7Z020} & 85k  & 0.23 
		\\  
		\midrule[0.1pt]		
		\scriptsize{Aydonat et al. \cite{Aydonat.2017}} & float & 16 & 16 & 0\% & 1382 & 45 & 303 & \scriptsize{Arria 10} & 1150\si{k} & 1.2
		\\
		\midrule[0.1pt]
		\scriptsize{Liang et al. \cite{Liang.2018}} & binary & 1* & 1* & \scriptsize{0.46-14.7\%} & 9396 & 150 & 26.2 & \scriptsize{Stratix V} & 695\si{k} & 13.52
		\\
		\midrule[0.1pt]
		\scriptsize{Chung et al. \cite{chung2018serving}} & float & 9 & 9 & 1-2\% & 39500 & 300 & 125 & \scriptsize{Stratix 10} & 2800\si{k} & 14.1
		\\
		\midrule[0.1pt]
		\multirow{2}{*}{\scriptsize{Ma et al. \cite{Ma.2018}}} & \multirow{2}{*}{fixed} & \multirow{2}{*}{16} & \multirow{2}{*}{16} & \multirow{2}{*}{2\%} & 348 & 150 & \multirow{2}{*}{N/A} & \scriptsize{Stratix V} & 622\si{k} & 0.56 \\
		& & & &  & 715 & 200 & & \scriptsize{Arria 10} & 1150\si{k} & 0.62 
		\\
		\midrule[0.1pt]
		\multirow{2}{*}{\scriptsize{Venieris et al. \cite{Venieris.2019}}} & \multirow{2}{*}{fixed} & \multirow{2}{*}{16} & \multirow{2}{*}{16} & \multirow{2}{*}{N/A} & 48.53 & \multirow{2}{*}{125} & $<$5 & \scriptsize{XC7Z020} & 85\si{k} & 0.57 
		\\
		& & & & & 155.81 & & $<$5 & \scriptsize{XC7Z045} & 350k  &  0.45\\
		\midrule[0.1pt]
		\scriptsize{Nguyen et al. \cite{Nguyen.2019}} & binary & 1* & 4-6* & 2.5\% & 1877 & 200 & 18.29 & \scriptsize{XC7VX485} & 485.7\si{k} & 3.36
		\\
		\midrule[0.1pt]
		\midrule
		\multicolumn{11}{l}{(a) compared to 32\si{bit} floating-point} \\
		\multicolumn{11}{l}{*Not applied to first and/or last layers.} \\
		\multicolumn{11}{l}{** Estimated value.} \\
		\bottomrule[1.5pt]
	\end{tabularx}
	\label{tab:FPGA-overview}
\end{table*}


\section{Comparison} \label{sec:comparision}
The decision of which neural network hardware accelerator is the most suitable depends on the application. For this reason, a separated comparison is done in terms of throughput, latency, accuracy, power, and usability. Since accuracy, throughput, and energy cost is a trade off \cite{Guo.24122017}. Based on that, the aim of the comparison is not to determine which platform is generally the best, but rather which platform should be selected for which application. \\
A distinction between throughput and latency is necessary since these parameters can differ greatly depending on the hardware accelerator and batch size. For real-time applications latency is the most important measure. Throughput is of minor interest. The opposite is true for all other applications. 

\subsection{Throughput} 
Due to the high clock frequency, the large die size, and the \gls{SIMD} architecture, which provides high parallelism, \glspl{GPU} can offer up to some Tera \gls{FLOPS} throughput \cite{Liang.2018,jouppi2017datacenter}. Due to the use of specialized \glspl{PE} and a lower bit width, \gls{ASIC} based implementations achieve higher throughput compared to general-purpose \glspl{GPU}. However, the introduction of specialized tensor processing units in \glspl{GPU}, \Cref{tab:ASIC-overview} and \Cref{tab:GPU-overview} show that \glspl{GPU} achieve a similar performance density compared to TPU using 8 bit fixed point quantization. Nevertheless, TPU is more power-efficient, and the die is manufactured with a larger feature size. \\ 
Compared to \glspl{GPU} and \glspl{ASIC}, \gls{FPGA} based implementation achieve a lower throughput. That comes from the fact that the variable routing of an \gls{FPGA} limits the bandwidth and is area intensive. 

\subsection{Latency} 
As mentioned before, latency is critical for real-time applications, which are implemented mainly in edge devices. In real-time applications, the ability to batch the input data is limited to meet the real-time requirements \cite{zhang2018dnnbuilder,chung2018serving}. \\
How efficient \glspl{GPU} process neural networks depends on the size of the matrices and thus on the batch size. For this reason, \glspl{GPU} lose performance in real-time applications, since the \gls{GEMM} based optimization is less efficient using a small batch size \cite{chung2018serving}. Therefore, \gls{ASIC} or \gls{FPGA} based implementations, which do not rely on \gls{GEMM} can achieve a better throughput to latency ratio. Chung et al. showed that \gls{FPGA} based hardware accelerators can outperform an Nvidia Tesla Xp \glspl{GPU} in real-time applications if fixed-point quantization is used \cite{chung2018serving}. Besides note that the higher design freedom of \glspl{ASIC} and \glspl{FPGA} enables more efficient processing of compact models. However, for a more complex data flow additional area is necessary. 

\subsection{Accuracy}
The accuracy of a \gls{DNN} mainly depends on the neural network structure, the dataset, and the training methods. These parameters are independent of the hardware accelerator. Many neural network hardware accelerators use quantization to increase performance density. However, that influences the accuracy of the required computations and can lead to an accuracy loss. Generally, for both floating-point and fixed-point quantization, using a bit width of 16 bits or higher does not lead to an accuracy loss. Otherwise, retraining can be used to recover the accuracy to minimize the accuracy loss due to quantization. Retraining can even lead to a performance gain for some not optimized neural network structures such as AlexNet  \cite{zhu2016trained}.\\
\Cref{tab:FPGA-overview} gives an overview of the used bit width and resulting accuracy loss. It shows that, if a minor accuracy loss is permitted, a more aggressive quantization method, such as binary weights in combination with 6-bit activations, can be used. That leads to an enormous gain in terms of throughput and latency \cite{Nguyen.2019}.
\subsection{Power}
Power consumption is a key-element for battery-powered embedded systems such as mobile devices, robots, etc. \cite{Liang.2018}. However, for data centers, power consumption is also of interest since energy and cooling costs make up a significant part of the running costs of a data center \cite{mohamed2019survey}. 
For this reason, \glspl{FPGA} and \glspl{ASIC} become more attractive for data centers since they offer advantages in the utilization of computation capacity and high energy efficiency \cite{chung2018serving,fowers2018configurable}. \Glspl{FPGA} usually achieve more than 10 times higher energy efficiency compared to \glspl{GPU} \cite{Liang.2018}. However, comparing \Cref{tab:GPU-overview} and \Cref{tab:FPGA-overview} shows that this is only true if the \gls{GPU} uses floating-point-32. \Cref{tab:ASIC-overview} visualizes that except for TPU, all other \gls{ASIC} implementations have a very high power efficiency. However, TPU has a better power efficiency compared to \glspl{GPU} and high throughput. Therefore, TPU is a favorable solution for data centers. \\
For power-critical embedded devices, \glspl{ASIC} or \glspl{FPGA} is a feasible choice. 

\subsection{Usability} 
For \Glspl{GPU} many frameworks already exist which offer a high-level API for creating, training, and testing neural networks (See \Cref{sec:gpu}). 
For using the tensor cores of a \gls{GPU} a C++ library called TensorRT is required, which is integrated into TensorFlow. It also includes a parser to import existing models from most other neural network training APIs. 
For this reason \glspl{GPU} provide the best usability compared to \glspl{ASIC} and \glspl{FPGA}. \\
However, TPU supports the high-level TensorFlow framework, which greatly increases usability. Since the TPU introduce by Jouppi et al. uses int8 quantization, training is not recommended \cite{jouppi2017datacenter}. However, next-generation TPUs support floating-point operations and training, therefore. Since other \gls{ASIC} implementations do not support high-level APIs they have poor usability.  
For \glspl{FPGA} some frameworks based on \gls{HLS} and OpenCL exist (See \Cref{sec:fpga}). Nevertheless, more in-depth knowledge of neural networks and hardware design is necessary for implementing an efficient \gls{FPGA} based hardware accelerator. Just like \glspl{ASIC}, \glspl{FPGA} usually do not support training. Compared to \glspl{ASIC}, \glspl{FPGA} have the advantage that they have a short design cycle, which is crucial in the field of neural networks because progress is fast \cite{wang2018survey}.

\section{Conclusions}
\label{sec_conclusions}
In the last years, the accuracy achieved by deep neural networks is increasing rapidly. For this reason, \glspl{DNN} are widely used for many \gls{AI} applications. However, the increasing accuracy bases on an exponential expansion of the parameter number, which leads to an exponential increase of computational load. This computational effort can hardly be handled by \glspl{CPU}. For this reason, optimized neural network hardware accelerators are required. Optimization can be achieved through algorithmic optimization, quantization, and \gls{DNN} compression. However, one has to be aware that the efficiency of optimization methods depends on the hardware accelerator's architecture, supported quantization, and dataflow. For this reason, a general assessment of the efficiency of individual techniques is not possible. Besides, quantization optimization and \gls{DNN} compression methods can lead to an accuracy loss. However,  post-optimization retraining can reduce accuracy loss. \\
\\\
Since general-purpose \glspl{GPU} provide a huge number of floating-point \glspl{ALU}, \glspl{GPU} are well-suited for the training of neural networks. However, the \gls{SIMD}/\gls{SIMT} architecture of \glspl{GPU} lead to many DRAM accesses. In combination with the use of power-hungry floating-point computations, the high number of DRAM accesses lead to the high power consumption of \glspl{GPU}. For this reason, in modern \glspl{GPU}, additional tensor processing units are added, which provide an improved data flow and a quantization down to int4. \\
In contrast to general-purpose \glspl{GPU}, \gls{ASIC}-based neural network hardware accelerators are extremely optimized for tensor computations. For this reason, \glspl{ASIC} provide both high throughput and a high power-efficiency. However, creating a fully custom neural network hardware accelerator requires a long design time and an in-depth knowledge of chip design and neural networks. Since the progress in the field of deep learning is fast, the time to market is essential. Additionally, a high-level API, which simplifies the implementation of a designed neural network, is required to enable all neural network architects to use the accelerator. \\
\Glspl{FPGA} offer a programmable logic circuit that enables a highly optimized design with comparable low design effort and initial costs. Additionally, some frameworks exist based on \gls{HLS}, which automatically generates an optimized hardware implementation based on the neural network design. Recent advances in sparsity and quantization enabled \glspl{FPGA} to achieve a throughput comparable to general-purpose \glspl{GPU} while having a higher power efficiency. Besides, \glspl{FPGA} offer the possibility to optimize the dataflow, to reduce the required memory bandwidth for sparse neural network structures. For this reason, \glspl{FPGA} are a promising candidate for the use in real-time edge systems. They also offer the advantage over \glspl{ASIC} that they are reprogrammable. Which is a considerable benefit due to the rapid progress in the field of deep learning.\\ 
\\\
It can be concluded that \glspl{FPGA} are a feasible choice for a real-time application with a limited power budget. \Glspl{ASIC} are well suited for the use in data centers since the high initial costs and the long time-to-market is an acceptable risk to achieve a higher performance and energy efficiency. For all other applications, \glspl{GPU} is the most feasible choice. 

\bibliographystyle{ACM-Reference-Format}
\bibliography{bibliography/bibliography}

\end{document}

%% file: Acronyms.tex
\newacronym{SNR}{SNR}{signal to noise ration}
\newacronym{FPGA}{FPGA}{field programmable gate array}
\newacronym{IP}{IP}{intellectual property}
\newacronym{ILA}{ILA}{integrated logic analyzer}
\newacronym{ADC}{ADC}{analog to digital converter}
\newacronym{CNN}{CNN}{convolutional neural network}
\newacronym{DMA}{DMA}{direct memory access}
\newacronym{PS}{PS}{processing system}
\newacronym{PL}{PL}{programmable logic}
\newacronym{BRAM}{BRAM}{block-RAM}
\newacronym{PE}{PE}{processing element}
\newacronym{LE}{LE}{logic element}
\newacronym{MAC}{MAC}{multiply and accumulate}
\newacronym{ASIC}{ASIC}{application specific integrated circuit}
\newacronym{GPU}{GPU}{graphics processing unit}
\newacronym{CPU}{CPU}{central processing unit}
\newacronym{RNN}{RNN}{recurrent neural network}
\newacronym{FC}{FC}{fully connected}
\newacronym{DNN}{DNN}{deep neural network}
\newacronym{ANN}{ANN}{artificial neural network}
\newacronym{SNN}{SNN}{spiking neural network}
\newacronym{BNN}{BNN}{binarized neural network}
\newacronym{TNN}{TNN}{ternary neural network}
\newacronym{ALU}{ALU}{arithmetic and logic unit}
\newacronym{AI}{AI}{artificial intelligence}
\newacronym{LUT}{LUT}{lookup table}
\newacronym{GEMM}{GEMM}{general matrix multiplication}
\newacronym{SIMD}{SIMD}{single instruction multiple data}
\newacronym{SIMT}{SIMT}{single instruction multiple threads}
\newacronym{OpenCL}{OpenCL}{open computer language}
\newacronym{FIR}{FIR}{finite impulse response}
\newacronym{TPU}{TPU}{tensor processing unit}
\newacronym{AOI}{AOI}{automated optical inspection}
\newacronym{MOI}{MOI}{manual optical inspection}
\newacronym{PCB}{PCB}{printed circuit board}
\newacronym{PCBA}{PCBA}{printed circuit board assembly}
\newacronym{MV}{MV}{machine vision}
\newacronym{DSP}{DSP}{digital signal processor}
\newacronym{IC}{IC}{integrated circuit}
\newacronym{HLS}{HLS}{high level synthesis}
\newacronym{RTL}{RTL}{register transfer level}
\newacronym{FFT}{FFT}{fast fourier transformation}
\newacronym{FF}{FF}{flip flop}
\newacronym{SP}{SP}{soft processor}
\newacronym{CLB}{CLB}{configurable logic block}
\newacronym{Relu}{ReLU}{rectified linear unit}
\newacronym{LSTM}{LSTM}{long short term memory}
\newacronym{GRU}{GRU}{gated recurrent unit}
\newacronym{IMC}{IMC}{in memory computation}
\newacronym{NFU}{NFU}{neural function unit}

%% file: Symbols.tex
\newglossaryentry{symb:lambda}{name={$\lambda$},description={Wavelength},type=symbolslist}
\newglossaryentry{symb:tau}{name={$\tau$},description={delay},type=symbolslist}
\newglossaryentry{symb:epsilon}{name={$\epsilon$},description={Measurement error},type=symbolslist}
\newglossaryentry{symb:frequency}{name={$f$},description={Frequency},type=symbolslist}
\newglossaryentry{symb:Gamma}{name={$\Gamma$},description={Reflection coefficient},type=symbolslist}
\newglossaryentry{symb:V}{name={$V$},description={Voltage},type=symbolslist}
\newglossaryentry{symb:I}{name={$I$},description={Current},type=symbolslist}
\newglossaryentry{symb:Z}{name={$Z$},description={Impedance},type=symbolslist}
\newglossaryentry{symb:G}{name={$G$},description={Gain},type=symbolslist}
\newglossaryentry{symb:R}{name={$R$},description={Resistance},type=symbolslist}
\newglossaryentry{symb:C}{name={$C$},description={Capacitance},type=symbolslist}
\newglossaryentry{symb:T}{name={$T$},description={Temperature},type=symbolslist}
\newglossaryentry{symb:Jpp}{name={$J_p_p$},description={Peak to peak jitter},type=symbolslist}
\newglossaryentry{symb:Jstd}{name={$J_\sigma$},description={Standard deviation of jitter},type=symbolslist}
\newglossaryentry{symb:Pre}{name={$\epsilon_\sigma$},description={Precision},type=symbolslist}
\newglossaryentry{symb:maxErr}{name={$\epsilon_m_a_x$},description={Maximum simulation error},type=symbolslist}
\newglossaryentry{symb:sps}{name={Sps},description={Samples per second},type=symbolslist}
\newglossaryentry{fps}{name={fps},description={frames per second},type=symbolslist}
\newglossaryentry{GOPS}{name={GOPS},description={giga operations per second},type=symbolslist}
\newglossaryentry{FLOPS}{name={FLOP/s},description={floating point operations per second},type=symbolslist}